\documentclass[10pt,twocolumn,letterpaper]{article}

\usepackage{wacv}      

\usepackage[accsupp]{axessibility}
\usepackage{graphicx}
\usepackage{amsmath}
\usepackage{amssymb}
\usepackage{multirow}
\usepackage{subcaption}
\usepackage{caption}
\usepackage{booktabs, etoolbox, soul}
\usepackage{tabularx}
\usepackage[pagebackref,breaklinks,colorlinks]{hyperref}
\usepackage{enumitem}
\usepackage[toc,page]{appendix}
\usepackage{algorithm}
\usepackage{algpseudocode}
\usepackage{listings}
\usepackage[frozencache=true,cachedir=minted-cache]{minted}
\usepackage{rotating}

\usepackage[capitalize]{cleveref}
\crefname{section}{Sec.}{Secs.}
\Crefname{section}{Section}{Sections}
\Crefname{table}{Table}{Tables}
\crefname{table}{Tab.}{Tabs.}

\newbool{showComments}
\booltrue{showComments}

\ifbool{showComments}{%
\newcommand{\cm}[1]{\sethlcolor{lime}\hl{[Cecilia: #1]}}
\newcommand{\lo}[1]{\sethlcolor{cyan}\hl{[Lorena: #1]}}
\newcommand{\cit}[1]{\sethlcolor{green}\hl{[Ian: #1]}}

\newcommand{\ds}[1]{\sethlcolor{yellow}\hl{[Dimitris: #1]}}
}{
\newcommand{\cm}[1]{}
\newcommand{\lo}[1]{}
\newcommand{\cit}[1]{}
\newcommand{\ds}[1]{}
}

\newcommand{\tool} {{\texttt{Kaizen}}\xspace}

\def\wacvPaperID{2406}
\def\confName{WACV}
\def\confYear{2024}

\begin{document}

\title{Kaizen: Practical Self-supervised Continual Learning \\ with Continual Fine-tuning}

\author{Chi Ian Tang$^{1,2}$, 
Lorena Qendro$^{1}$, Dimitris Spathis$^{1}$, Fahim Kawsar$^1$,\\Cecilia Mascolo$^2$, and Akhil Mathur$^1$\\ 
\\
$^1$Nokia Bell Labs, UK\\
$^2$University of Cambridge, UK\\
$^1${\tt\small \{ian.tang, lorena.qendro, dimitrios.spathis, fahim.kawsar\}@nokia-bell-labs.com} \\
$^2${\tt\small \{cit27, cm542\}@cam.ac.uk} \\
}

\maketitle

\begin{abstract}

Self-supervised learning (SSL) has shown remarkable performance in computer vision tasks when trained offline. However, in a Continual Learning (CL) scenario where new data is introduced progressively, models still suffer from catastrophic forgetting. Retraining a model from scratch to adapt to newly generated data is time-consuming and inefficient. Previous approaches suggested re-purposing self-supervised objectives with knowledge distillation to mitigate forgetting across tasks, assuming that labels from \textit{all} tasks are available during fine-tuning. In this paper, we generalize self-supervised continual learning in a practical setting where available labels can be leveraged in any step of the SSL process. With an increasing number of continual tasks, this offers more flexibility in the pre-training and fine-tuning phases. With \tool\footnote{The code for this work is available at \\ \url{https://github.com/dr-bell/kaizen}}, we introduce a training architecture that is able to mitigate catastrophic forgetting for both the feature extractor and classifier with a carefully designed loss function. By using a set of comprehensive evaluation metrics reflecting different aspects of continual learning, we demonstrated that \tool significantly outperforms previous SSL models in competitive vision benchmarks, with up to 16.5\% accuracy improvement on split CIFAR-100. \tool is able to balance the trade-off between knowledge retention and learning from new data with an end-to-end model, paving the way for practical deployment of continual learning systems.

\end{abstract}
\vspace*{-5mm}
\section{Introduction}

While traditional machine learning algorithms perform well on tasks they have been exposed to during training, they are unable to adapt to new concepts, classes, or data introduced after training over time. This is particularly important in vision tasks, where deployed models such as those running on phones or CCTV cameras need to adapt to new information. Currently, model updates require, first, offline re-training on a central server and then distributing the updated model to the end users. This procedure greatly affects the usability and user experience since new functionalities such as recognizing new people in a photo album app~\cite{meng2021magface}, will not be included until a large amount of data has been collected for re-training. Given these limitations, a system that can adapt to new data and learn continually would be highly desirable.

\begin{figure}[t]
\begin{center}
    \begin{subfigure}{\linewidth}
        \centering
        \includegraphics[width=0.88\linewidth]{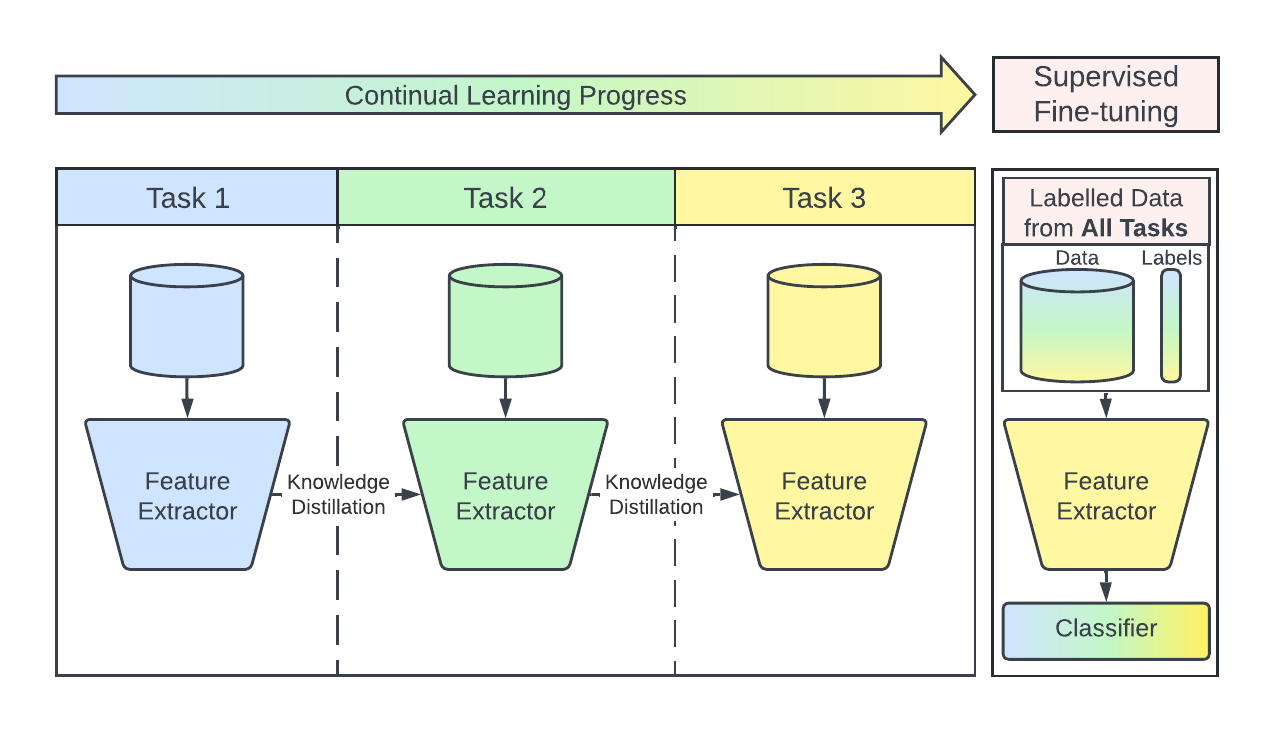}
        \vspace{-0.3cm}
        \caption{Self-supervised Continual Learning}
        \label{fig:sscl}
    \end{subfigure}
    \begin{subfigure}{\linewidth}
        \centering
        \includegraphics[width=0.88\linewidth]{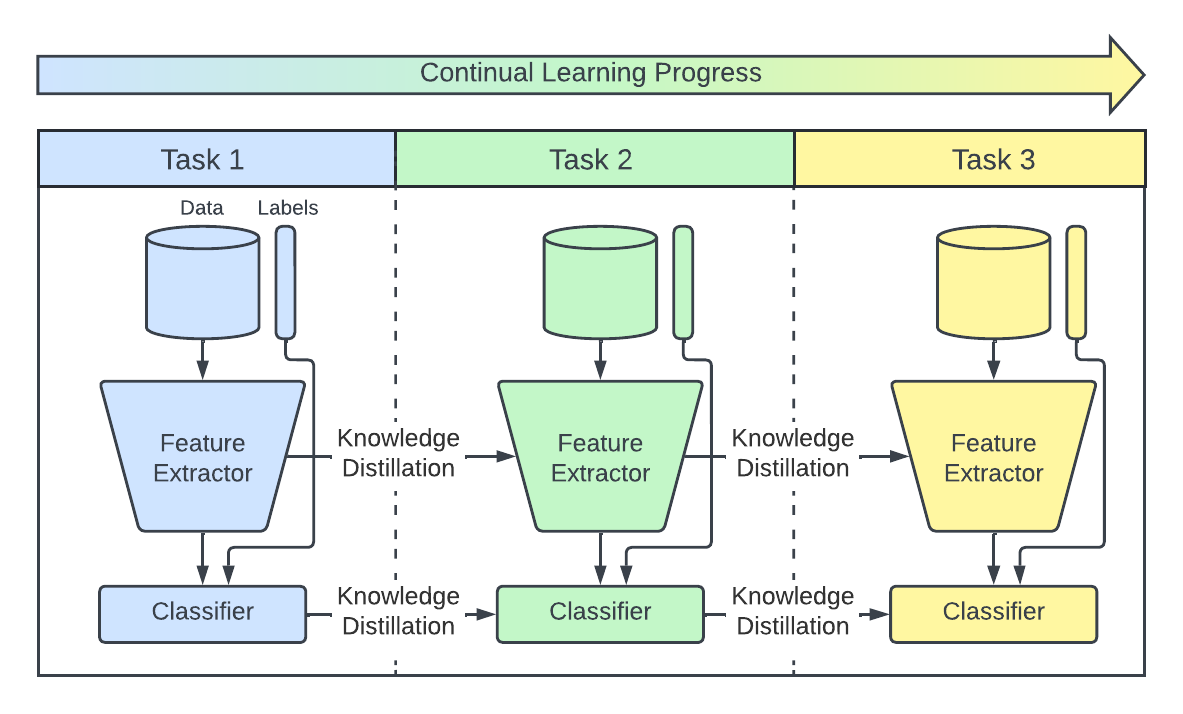}
        \caption{\tool: Self-supervised Continual Learning with Continual Fine-tuning}
        \label{fig:sscl_ft}
    \end{subfigure}%
\end{center}
   \caption{ \textbf{Self-supervised Continual Learning vs Continual Fine-tuning}. Existing approaches (a) wait until the end of the continual process to fine-tune, \tool (b) leverages distillation across both the feature extraction and fine-tuning steps for each task. 
   }
\end{figure}

Continual learning (CL) methods aim to learn continually from an ever-changing stream of data, acquiring new knowledge while retaining performance on previous tasks~\cite{van2019three}. The main difference to conventional deep learning is the data assumption: rather than having all training data available at once, data with different distributions (classes or domains) arrive over time. In practice, data can be only temporarily available for training because of storage, computing, and memory limitations as well as privacy concerns. Data availability comes also with the cost of labelling samples for supervised learning or fine-tuning, since only a limited amount of data can be annotated at a time.

Self-supervised Learning (SSL) techniques solve the data availability--annotation challenge even outperforming supervised methods in difficult vision tasks. SSL models learn meaningful representations that can be transferred to various downstream applications ~\cite{chen2020improved,bardes2022vicreg,grill2020bootstrap,chen2020simple}. However, they usually operate under strict data assumptions where training is conducted on large aggregated datasets. These assumptions no longer hold when new unlabelled data is introduced over time and there is not a \textit{single} snapshot of the dataset to train on. As a result, SSL models should adapt to the temporary availability of both unlabelled and labelled data. On the other hand, prior works in CL almost exclusively focus on supervised setups using ideas such as regularization, replay, or parameter-isolation \cite{serra2018overcoming, kirkpatrick2017overcoming, shin2017continual}. Our aim is to leverage both SSL and CL paradigms, therefore here we focus on Continual Self-Supervised Learning (CSSL), a still under-explored area. Some recent works \cite{fini2022self, lin2022continual} have started looking at CSSL where models learn from a stream of unlabelled data, however, these methods focus on training a strong feature extractor assuming that labelled data from \textit{all} tasks\footnote{A task in this context refers to a particular distribution of data, which can come from different sets of classes (for classification problems) or different domains.} are available for fine-tuning. We argue that this violates the core CL assumption where limited to no data from earlier tasks is available. A more practical setting would involve leveraging both unlabelled and labelled data \textit{within each task's training round}.

\begin{figure}
\begin{center}
   \includegraphics[width=0.85\linewidth] {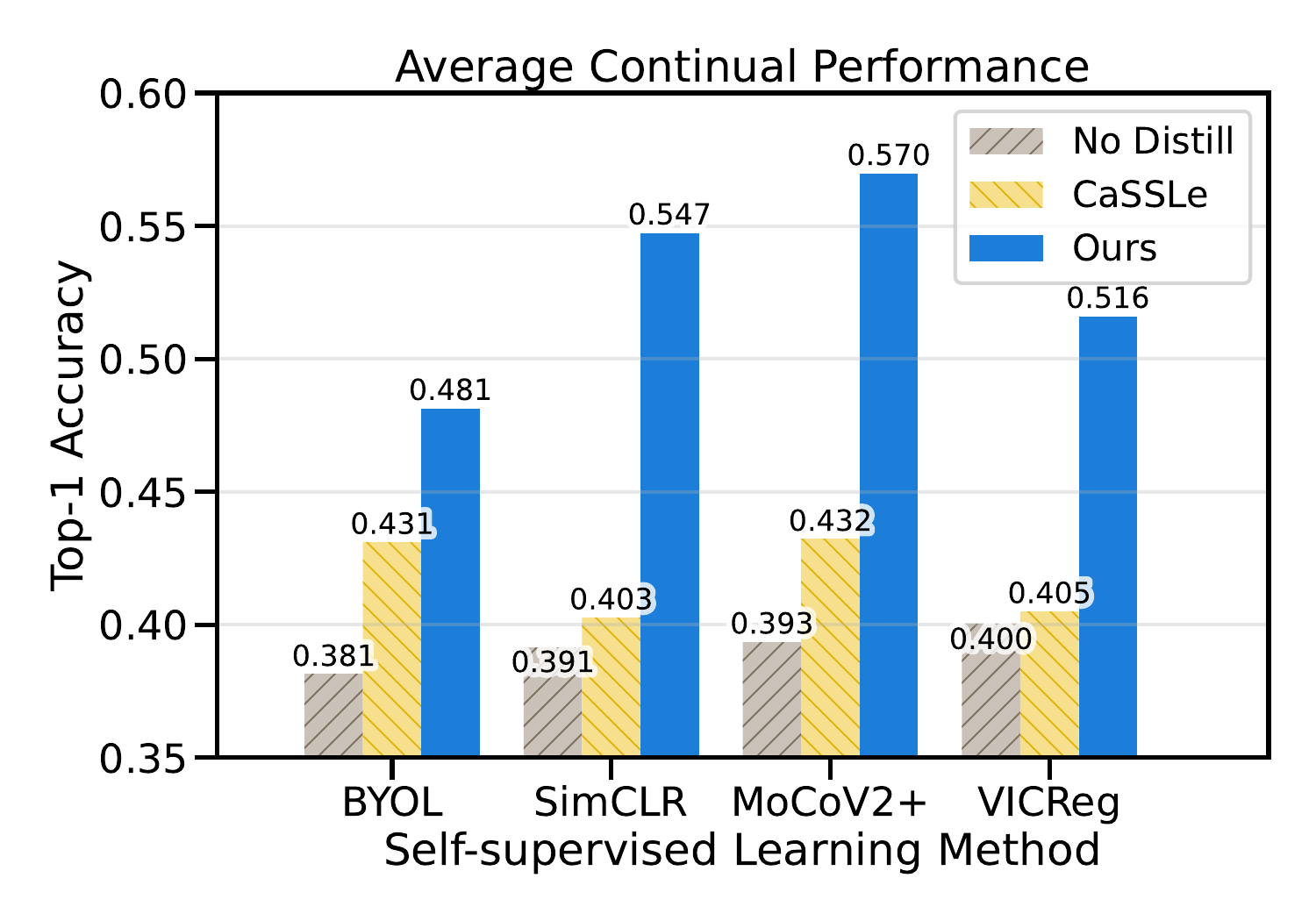}
    
   \includegraphics[width=0.85\linewidth]{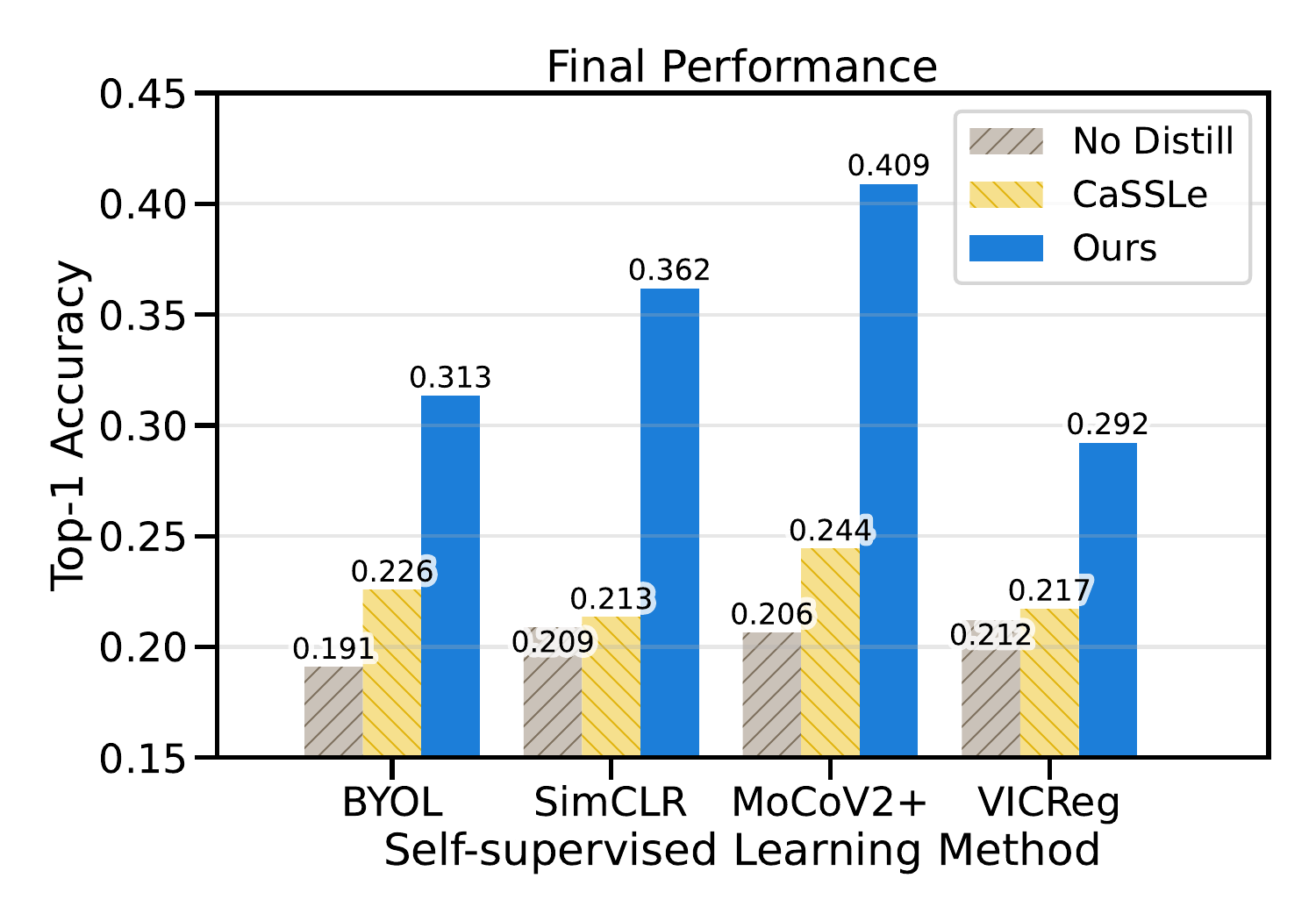}
   \vspace{-1cm}
   
\end{center}
   \caption{ \textbf{Performance comparison}. Models trained using different self-supervised learning methods and knowledge distillation strategies on class-incremental CIFAR-100. The top figure shows the average performance across the entire continual learning process, while the bottom figure shows the performance in the final evaluation. 
   }
   \label{fig:general_performance_comparison}
\end{figure}

To achieve that, we introduce \tool\footnote{inspired by the East Asian concept of ``continuous improvement"}, a new CSSL architecture designed to work under a more practical data assumption, where end-to-end training is performed using the continuous arrival of unlabelled and labelled data. This architecture strikes a balance between (i) training the feature extractor and fine-tuning, and (ii) combating catastrophic forgetting and learning from new data by employing a novel loss function that is specifically designed for these objectives. 

We demonstrate that \tool outperforms previous works that focus on self-supervised continual learning in terms of their overall performance and the ability to retain knowledge after training on the final task, and more importantly after training on every task, by up to 14.4\% in accuracy (in absolute terms) and 25.4\% reduction in forgetting. Our main contributions are:

\begin{enumerate}[leftmargin=*]
    \item \textbf{Practical Continual Learning framework.} We propose \tool, a continual learning framework that can be deployed at any point during the continual learning process with a functional classifier. It leverages both unlabelled and labelled data in training the feature extractor and classifier instead of only using one type of data, in a carefully designed loss function and distillation mechanism, thus allowing higher flexibility in terms of storage requirements and in accommodating privacy concerns, which is important in real-world applications.
    \item \textbf{Evaluation setup reflecting the real world.} We propose a novel evaluation setup with a range of evaluation metrics that closely reflect how models perform in the real world, where there is a focus on the performance of the model over the entire continual learning process, instead of only the final model.
    \item \textbf{Extensive empirical analysis.} We extensively evaluate \tool in various settings and vision benchmarks with state-of-the-art self-supervised learning techniques. We show that \tool is robust to catastrophic forgetting in classification tasks when encountering new data, while maintaining the quality of the feature extractor.
\end{enumerate}

\section{Related Work}

\noindent \textbf{Continual Learning.} Traditional deep learning techniques suffer from catastrophic forgetting~\cite{kirkpatrick2017overcoming}, a phenomenon where models tend to forget what has been previously learned by overriding it with new data. Continual learning methods address this issue by striking a balance between the stability (ability to retain knowledge) and plasticity (ability to learn new concepts) of deep learning models. Broadly, continual learning techniques can be categorized into three types~\cite{de2021continual}: replay-based~\cite{rebuffi2017icarl, ostapenko2019learning, chaudhry2018efficient, buzzega2020dark}, where a small subset of previously seen data is retained and replayed to the model at every stage, regularisation-based~\cite{kirkpatrick2017overcoming, li2017learning, shin2017continual, wu2019large}, in which regularisation objectives are used to guide the model to retain old knowledge, and parameter-isolation methods~\cite{rusu2016progressive, serra2018overcoming}, where separate regions/neurons of the models are dedicated to each task and remain frozen during further training. {\em Nevertheless, these techniques mostly focus on supervised learning setups, without dealing with label scarcity which is a common issue in real-world scenarios}.

\noindent \textbf{Self-supervised Learning (SSL).} SSL methods use unlabelled data with the goal of training more generalizable models. Recent techniques which involve bringing the representations of correlated views (e.g. transformed versions) of a single data point (e.g., image) closer to each other have proven powerful where the learned representations of the feature extractor achieve  performance on par with those trained in a supervised manner~\cite{wu2018unsupervised, chen2020simple, grill2020bootstrap}. These techniques include contrastive methods such as  SimCLR~\cite{chen2020simple}, similarity maximization methods such as BYOL~\cite{grill2020bootstrap}, momentum-based methods such as MoCo~\cite{he2020momentum, chen2020improved}, and reduction-based methods like VicReg~\cite{bardes2022vicreg}. However, in continual learning settings, these methods still suffer from catastrophic forgetting~\cite{fini2022self, de2021continual}. Furthermore, they require substantial amounts of unlabelled data available \textit{all at once} to work well, which is not practical in scenarios considering streams of unlabelled data.

\noindent \textbf{Continual Self-supervised Learning (CSSL).}
CSSL approaches operate on unlabelled data while dealing with catastrophic forgetting. Earlier techniques focus either on self-supervised pre-training to later apply supervised continual learning~\cite{gallardo2021self, caccia2022special} or extend the contrastive SSL paradigm to CL, too~\cite{cha2021co2l, madaan2021rethinking}. However, these techniques have a narrow focus as they are tailored to specific SSL architectures. Few recent works~\cite{fini2022self, de2021continual} started looking at general frameworks for unsupervised continual learning, where the model learns continually from a stream of unlabelled data using a combination of unsupervised learning techniques and knowledge retention mechanisms. Nevertheless, these methods only solve half of the problem in CL -- they focus on training a strong feature extractor continually with unlabelled data and assume that all labelled data are stored and available for fine-tuning after the final CL step. \emph{This violates the data assumption in continual learning where labelled data is only temporarily available, and the classifier should also learn continually.}  \tool puts forward a general architecture in which both the feature extraction and the fine-tuning are performed continually from a stream of both large unlabelled and small labelled data.

\section{Method}
\subsection{Background}

\noindent \textbf{Continual Learning (CL)}.
In this work, we adopt the notion of \emph{task-incremental} learning \cite{de2021continual}, in which the model sees the training data for one task at a time. We assume data $(\mathcal{X}^{(t)},\mathcal{Y}^{(t)})$, which is randomly drawn from distributions $P(\mathcal{X}^{(t)})$ and $P(\mathcal{Y}^{(t)})$ for task $t$, is available for the model to learn from, with the goal to optimize for all seen tasks while having little to no access to data $(\mathcal{X}^{(t')},\mathcal{Y}^{(t')})$ from previous tasks $t' < t$ \cite{de2021continual}.
Here we focus on the \emph{class-incremental} learning \cite{hsu2018re, de2021continual} setting where each task contains an exclusive subset of classes in a dataset, and thereby ${P(\mathcal{X}^{(i)})\neq P(\mathcal{X}^{(j)})}$ and ${P(\mathcal{Y}^{(i)})\neq P(\mathcal{Y}^{(j)})}$ if $i \neq j$, \emph{without} the task label $t$ provided to the model during inference or evaluation.

\vspace{0.1in}

\noindent \textbf{SSL for visual representations}. 
Siamese representation learning or contrastive learning proposed in recent works \cite{chen2020simple, he2020momentum, chen2020improved, grill2020bootstrap, caron2020unsupervised, chen2021exploring, zbontar2021barlow, caron2021emerging, bardes2022vicreg} have demonstrated strong supervisory signal for a model to learn to extract distinctive features from large unlabelled datasets. The main idea of these methods involves passing two stochastically augmented views $x_1$ and $x_2$ of an input sample $x$ through the same feature extractor (or one through the feature extractor and another through an exponentially updated momentum feature extractor in some works \cite{grill2020bootstrap}), and optimizing using a loss function such that the extracted features are similar, with different mechanisms such as gradient stopping \cite{grill2020bootstrap, chen2021exploring} or a clustering task \cite{caron2021emerging} to avoid collapse. The stochastic augmentations are designed such that the output images encode the semantics that the model learns to extract, in which they still contain almost the same meaning but the appearance might be different.

\subsection{Kaizen: Practical continual learner that balances self-supervised learning and fine-tuning} \label{subsection:method}

\begin{figure}
    \centering
    \includegraphics[width=0.8\linewidth]{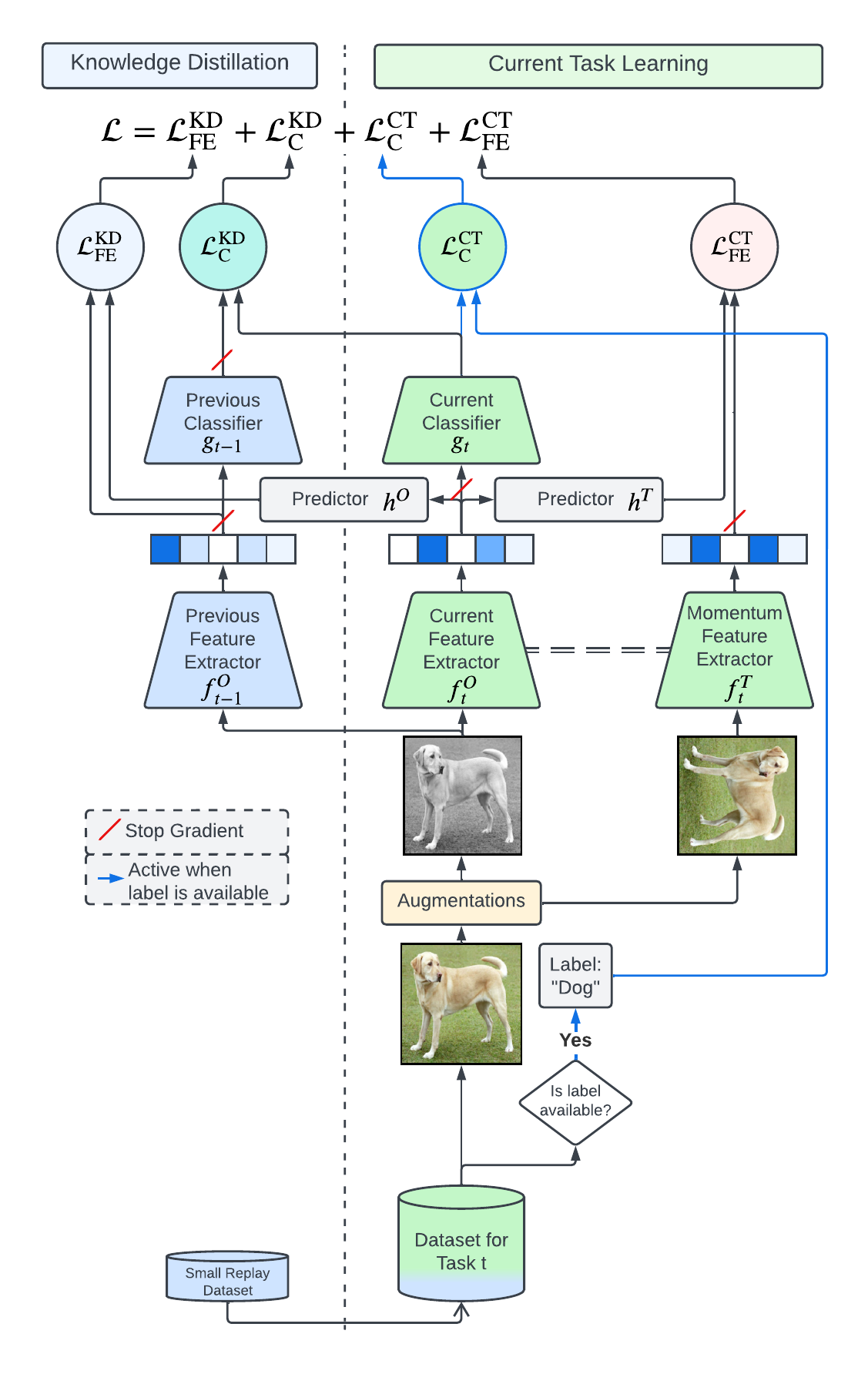}
    \caption{\textbf{Overview of the \tool framework}. \tool balances knowledge distillation and current-task learning in an end-to-end manner through a joint loss function. Available SSL methods can be used in training the feature extractors alongside knowledge distillation, while the classifiers are trained  on both unlabelled and labelled data through knowledge distillation and fine-tuning.}
    \label{fig:overview}
\end{figure}

In order to continually learn from unlabelled data while maintaining a functional classifier, the \tool framework (as illustrated in Figure~\ref{fig:overview}) is designed to balance different objectives including knowledge retention, self-supervised learning from unlabelled data, and supervised learning for classification. It consists of two main components: knowledge distillation and current task learning, one of each for the feature extractor and for the classifier.

\noindent \textbf{Current task learning.}
This component trains the feature extractor and the classifier using the data for the current task (see the right half of Figure~\ref{fig:overview}). In order to allow the model to learn from both unlabelled and labelled data, the feature extractor is trained on the current task using one of the self-supervised learning methods described above. Stochastic augmentation functions are applied to the input image which produces two different but correlated views of the image. One of these images is passed to the Current Feature Extractor $f_t^O$, while the other to the Momentum Feature Extractor $f_t^T$ (or the same Current Feature Extractor depending on the SSL method), to obtain two different embeddings. The embedding from the Current Feature Extractor is passed to an additional, shallow neural network called the predictor $h^T$. The discrepancy between the embedding from the Momentum Feature Extractor and that from the predictor will then be reduced using the corresponding self-supervised loss function $\mathcal{L}^{\mathrm{CT}}_{\mathrm{FE}}$. The use of an additional predictor here is to allow the feature extractor to be flexible so that it does not have to produce the same embedding for the two augmented views, but should contain the same amount of information.

If the label for the image is available, we train the classifier in a supervised manner. The embedding obtained from the Current Feature Extractor will be fed to the classifier network $g_t$ to obtain class probabilities after softmax activation. Categorical cross-entropy loss $\mathcal{L}^{\mathrm{CT}}_{\mathrm{C}}$ is used for training the classifier. Note that the gradient updates do not back-propagate to the feature extractor. This is to allow the feature extractor to focus on extracting distinctive features from self-supervision, instead of specialising in the current classification task. This is important to ensure that the feature extractor remains general throughout the continual learning process, where the class distribution might shift from one task to another.

\noindent \textbf{Knowledge distillation.}
Apart from the first task, the model is required to retain knowledge learned from previous tasks while learning from the new data. To achieve this, we make a frozen copy of the trained model ($f_{t-1}^O$ and $g_{t-1}$) from the previous task for knowledge distillation before we start training (Figure~\ref{fig:overview} (left)).
For the feature extractor, we adopt a scheme similar to CaSSLe \cite{fini2022self}, where we mirror what we have done for the current task learning, by using an SSL method but with the Momentum Feature Extractor replaced by the feature extractor from the previous step $f_{t-1}^O$. The augmented image that was passed through the Online Feature Extractor $f_t^O$ will also be fed to the Previous Feature Extractor. Similarly, the discrepancy between the embedding from the previous and the current feature extractor (after passing through a different predictor $h^O$) is reduced using the corresponding self-supervised loss function $\mathcal{L}^{\mathrm{SSL}}$. The Previous Feature Extractor is frozen during training and gradients updates will not be applied to it.

Similar to the feature extractor, knowledge distillation is also performed for the classifier. The predictions from the Current Classifier $g_t$ are made to be similar to the predictions (or soft labels) from the Previous Classifier ($g_{t-1}$) using the categorical cross-entropy loss. Note that unlike in current task learning, knowledge distillation for the classifier is active regardless of the presence of a label given the input image.

\textbf{Memory replay.}
One additional mechanism that our method employs, is memory replay \cite{rebuffi2017icarl, isele2018selective, rolnick2019experience}, where a small subset of actual samples or representative samples from previous tasks are kept and replayed during training. This has been shown to be crucial in maintaining model performance and combating catastrophic forgetting in task-label-free methods \cite{de2021continual}, where the task label is not provided to the model during inference.

\textbf{Overall Framework.}
The overall loss function (Equation \ref{eq:loss}) consists of four components: the loss for knowledge distillation of the feature extractor $\mathcal{L}^{\mathrm{KD}}_{\mathrm{FE}}$, the loss for knowledge distillation of the classifier $\mathcal{L}^{\mathrm{KD}}_{\mathrm{C}}$, the cross-entropy loss of the classification task $\mathcal{L}^{\mathrm{CT}}_{\mathrm{C}}$, and the loss for self-supervised learning $\mathcal{L}^{\mathrm{CT}}_{\mathrm{FE}}$. These losses balance different learning objectives to ensure that the model, including the feature extractor and the classifier, is able to learn from new data while retaining knowledge from previous tasks.
\begin{align}
\begin{aligned}
        \mathcal{L} = &\mathcal{L}^{\mathrm{KD}}_{\mathrm{FE}} + \mathcal{L}^{\mathrm{KD}}_{\mathrm{C}} + \mathcal{L}^{\mathrm{CT}}_{\mathrm{C}} + \mathcal{L}^{\mathrm{CT}}_{\mathrm{FE}} \\
        = &\mathcal{L}_{\mathrm{SSL}}\big(f_{t-1}^O(x_1), h^O\big(f_t^O(x_1)\big)\big) + \\
        &\mathcal{L}_{\mathrm{CE}}\big(g_{t-1}\big(f_{t-1}^O(x_1)\big), g_t\big(f_t^O(x_1)\big)\big) + \\
        &\mathcal{L}_{\mathrm{CE}}\big(g_{t}\big(f_{t}^O(x_1)\big), y\big) + \\
        &\mathcal{L}_{\mathrm{SSL}}\big(f_{t}^T(x_2)\big), h^T\big(f_t^O(x_1)\big)\big)
\end{aligned}
\label{eq:loss}
\end{align}

Here the $\mathcal{L}_{\mathrm{CE}}$ refers to the categorical cross-entropy loss, and $\mathcal{L}_{\mathrm{SSL}}$ refers to the self-supervised loss for the particular self-supervised learning method.

\section{Evaluation for Continual Learning}
In this section, we first discuss the evaluation protocol and metrics adopted in this work and then present our experimental setup.

\subsection{Self-supervised Continual Learning vs \\Continual Fine-tuning}
Pure self-supervised continual learning focuses on training a strong feature extractor from data with changing distributions, and after learning from the final set of data, a classifier is trained with labelled data (see Figure \ref{fig:sscl}). Although the feature extractor is learning continually, this only solves half of the challenge in combating catastrophic forgetting. This is because if we want a classifier that is able to perform classification on all seen tasks, labelled data from all tasks must be retained or re-collected for training. In the evaluation, we show that if we limit the amount of labelled data from earlier tasks to a small subset, the performance suffers significantly.

We argue that in a more practical scenario, the classifier should be trained together with the feature extractor (see Figure \ref{fig:sscl_ft}) because the labelled data is more likely to be available while training for a particular task, rather than after the feature extractor has been trained. This would require extra consideration with regard to knowledge distillation for the classifier on top of the feature extractor. Our approach is designed to address this issue.

\subsection{Evaluation metrics}
\label{subsection:evaluationi_metrics}
\begin{figure}[t]
    \centering
    \includegraphics[width=0.7\linewidth]{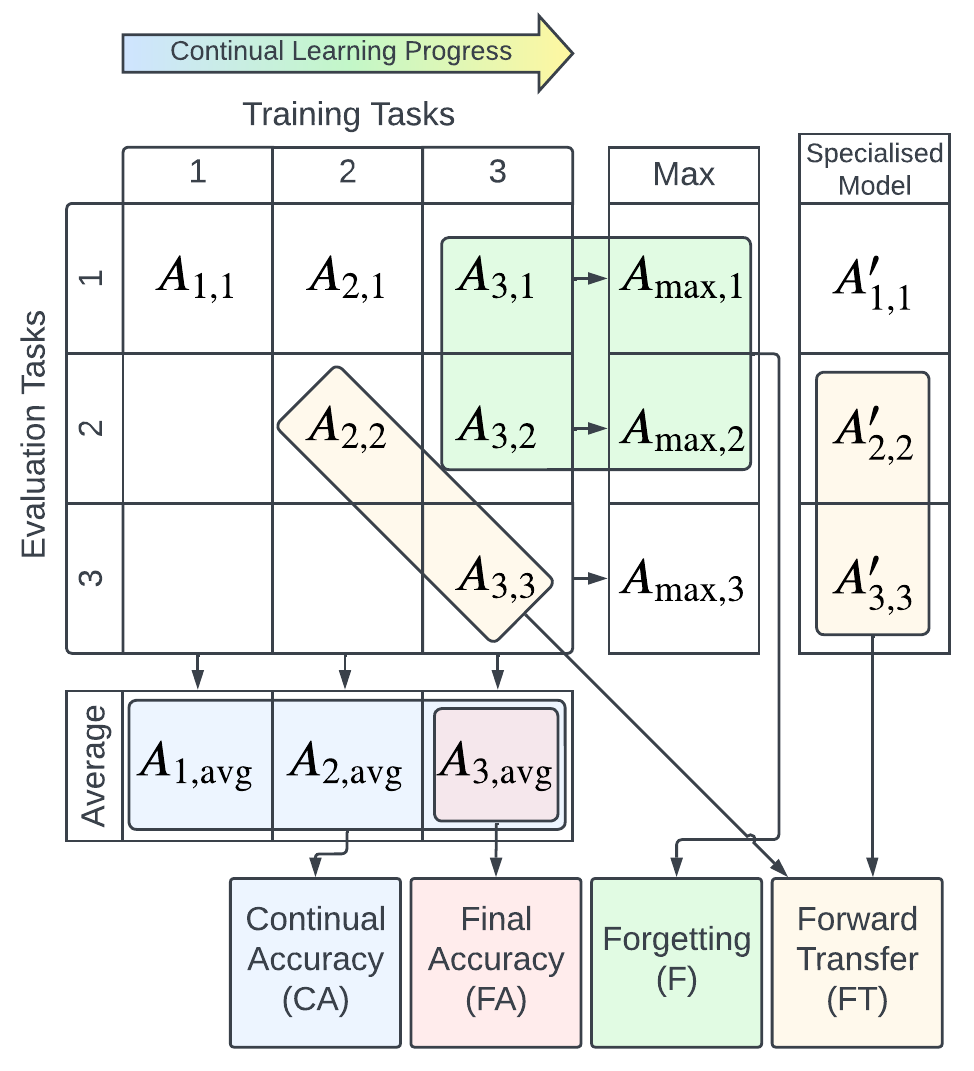}
    \caption{\textbf{Calculation of the evaluation metrics}. Illustration of our practical evaluation setup with regards to metrics and tasks used in each calculation.}
    \label{fig:metrics}
\end{figure}

In order to compare the performance of different methods fairly, we define the following evaluation metrics (see Figure~\ref{fig:metrics}) that reflect different aspects of a continual learning framework, taking into account metrics defined in previous works~\cite{de2021continual, diaz2018don, isele2018selective}. Here we denote the accuracy of a model on a particular task $k$ (averaged across all classes in that task) after seeing the last sample from task $t$ as $A_{t,k}$, and we assume that there are $T$ tasks in total.

\noindent \textbf{Final Accuracy (FA)} defined as $ FA = \frac{1}{T} \sum^T_{i=1} {A_{T,i}}$ refers to the average model accuracy across all tasks, after it has been trained on the last task $T$.

\noindent \textbf{Continual Accuracy (CA)}, instead of only looking at the performance of the model at the final step, looks at the performance throughout the continual learning process, where a model might be deployed before the last task and be later re-trained. We define this to be the average accuracy of the model after being trained on each task $CA = \frac{1}{T} \sum^T_{i=1} \big( \frac{1}{i}\sum^i_{j=1}{A_{j,i}} \big) $.
    
\noindent \textbf{Forgetting (F)} measures the extent of the performance a model has lost after training on new tasks. We calculate this by taking the difference between the highest performance of the model on a given task, and that of the model at the final step, excluding the final task: $
 F = \frac{1}{T-1} \sum^{T-1}_{i=1} \big({A_{max, i} - A_{T,i}} \big)$, where $A_{max, i} = \mathrm{max}_{t\in \{1, ..., T\}} A_{t, i}$.
    
\noindent \textbf{Forward Transfer (FT)} refers to the use of previously acquired knowledge when learning new tasks. Here, we define it with model deployment considerations in mind: we take the difference between the performance of the model on task $k$ after seeing data from all tasks $i \leq k$, compared to a model which has only seen data of task $k$. We denote the accuracy of the model which is only trained on task $k$ as $A'_{k, k}$. FT is thus calculated by: $
    FT = \frac{1}{T-1} \sum^{T}_{i=2} \big({A_{i, i} - A'_{i,i}} \big)$.
A positive value would indicate that the training on previous tasks helped the model learn the new task. Note that this is a desired property that is hard to achieve.

This evaluation framework is set to facilitate comparisons of different properties of CL models in real-world applications and balance trade-offs between different properties such as forgetting and forward transfer. These metrics should provide guidance for which method to select depending on desired use cases.

\subsection{Experimental setup}

\noindent \textbf{Datasets}. We performed our evaluation using 2 datasets: CIFAR-100 \cite{krizhevsky2009learning}, an object recognition dataset with 60,000 32x32 images evenly distributed across 100 classes, and ImageNet100 \cite{tian2020contrastive}, also an object recognition dataset which is a 100-class subset of the original ImageNet dataset \cite{deng2009imagenet} and consists of 130,000 224x224 images spread across 1000 classes. In our experiments, we randomly split these datasets into tasks with an equal number of classes. For example, for 5 tasks, we have 20 classes in each task.

\noindent \textbf{SSL methods}. Since we do not modify the self-supervised learning component from its original formulation~\cite{chen2020improved,bardes2022vicreg,grill2020bootstrap,chen2020simple} and the CaSSLe adaptation~\cite{fini2022self}, our method is compatible with existing self-supervised learning methods. We selected the contrastive-based methods  SimCLR~\cite{chen2020simple} and MoCoV2+~\cite{he2020momentum, chen2020improved}, the asymmetric-model-based method BYOL~\cite{grill2020bootstrap} and the cross-correlation-based method VICReg~\cite{bardes2022vicreg} as the backbone SSL method for training. Our aim is to investigate whether our proposed architecture generalizes across different self-supervised learning methods and identify limitations.

\noindent \textbf{Hyperparameters}. We build upon the Pytorch implementation introduced by \cite{fini2022self}. For each task, we train the model for 500 epochs for CIFAR-100 with 5 tasks, 250 epochs for CIFAR-100 with 20 tasks, and 200 epochs for ImageNet100. The batch size is 256 for CIFAR-100 and 128 for ImageNet100. We keep the amount of data replayed to the model during continual learning (for our method) and during classifier training (for other baselines) to be 1\% of the original data as this offered the best tradeoff between accuracy and amount of labelled data used. ResNet18~\cite{he2016deep} is used as the architecture for the feature extractor, while the classifier consists of one fully-connected layer of 1000 units and another one with the number of output classes, which is 100. LARS~\cite{you2017large} is used for large batch training. A weighting factor of 2 is used for the knowledge distillation loss for the classifier ($\mathcal{L}^{\mathrm{KD}}_{\mathrm{C}}$ in Equation~\ref{eq:loss}). We keep all other hyperparameters unmodified. 
In the supplementary material, we provide a more detailed discussion of the hyperparameters and the source code for reproducibility.

\noindent \textbf{Baselines}.\label{subsection:baselines} We compare our framework against the state-of-the-art CSSL pipeline, CaSSLe~\cite{fini2022self}, and the \emph{No distill} setup, where no extra measures are taken to mitigate catastrophic forgetting, and the entire model is fine-tuned from task to task. Since the baseline methods are pure self-supervised pre-training methods, the classifier is trained after the feature extractor is fully trained using the task data. To ensure a fair comparison, these classifiers are trained with memory replay enabled: the same subset of data that our model is trained on, is also available for these classifiers to train on.

\section{Results}

\subsection{Performance comparison against CSSL}
As discussed in section \ref{subsection:baselines}, we focus on the comparison of our \tool against CaSSLe and the \emph{No distill} setup. Figure~\ref{fig:general_performance_comparison} compares the Continual Accuracy and Final Accuracy of these three methods with different SSL models for the feature extractor, on the CIFAR-100 dataset split into 5 tasks of equal sizes. \tool outperforms the other two baselines irrespective of the evaluation metric or the SSL model, achieving the highest continual accuracy at 0.570 and final accuracy at 0.409 using MoCoV2+, outperforming CaSSLe by 0.138 and 0.165 respectively. It is important to note that the data availability is kept the same across all methods, where at each task, every method has access to data of the current task and 1\% of replay data from previous tasks. Our method outperforms the rest by incorporating knowledge distillation and fine-tuning into the pipeline, instead of training the classifier training in the end. This validates our hypothesis and shows that \tool is overall effective in retaining knowledge from previous tasks, by performing well at both the final step and throughout the continual learning process. It is interesting to note that MoCoV2+ is the most effective of the four SSL methods; this could potentially be attributed to the use of the Memory Buffer of representations, which acts as a knowledge retention mechanism that enables gradual learning.

\subsection{Performance variation across time}

\begin{figure}[t]
    \centering
    \includegraphics[width=0.49 \linewidth ]{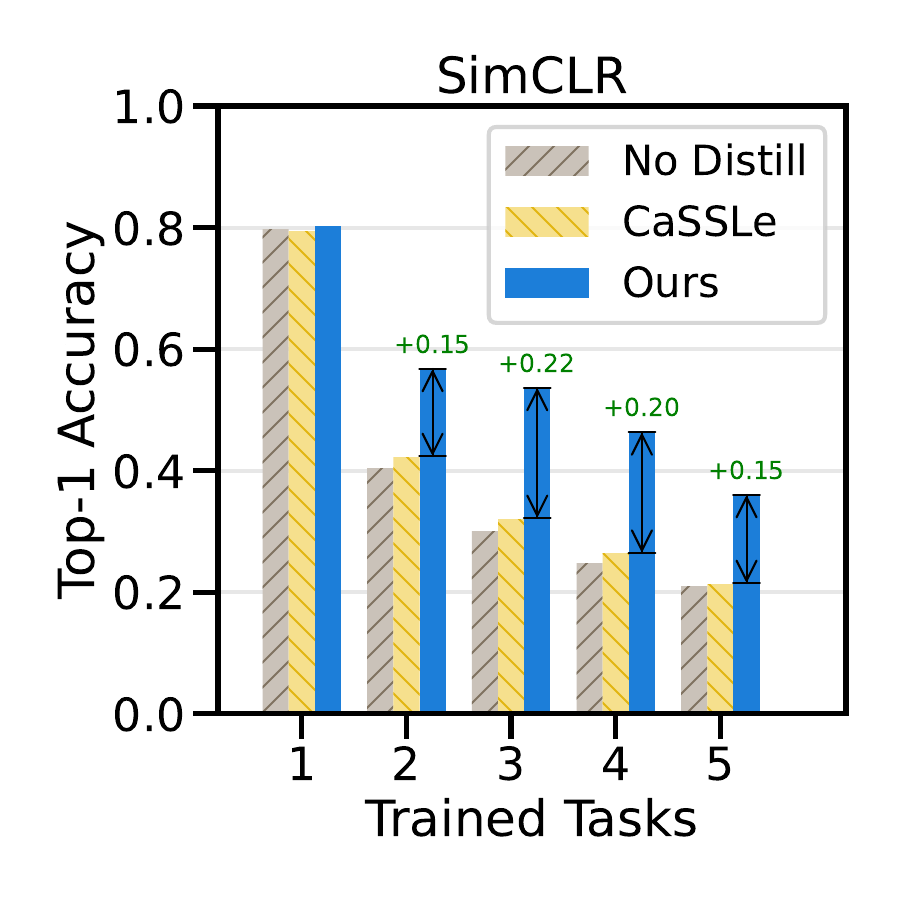}
    \includegraphics[width=0.49 \linewidth ]{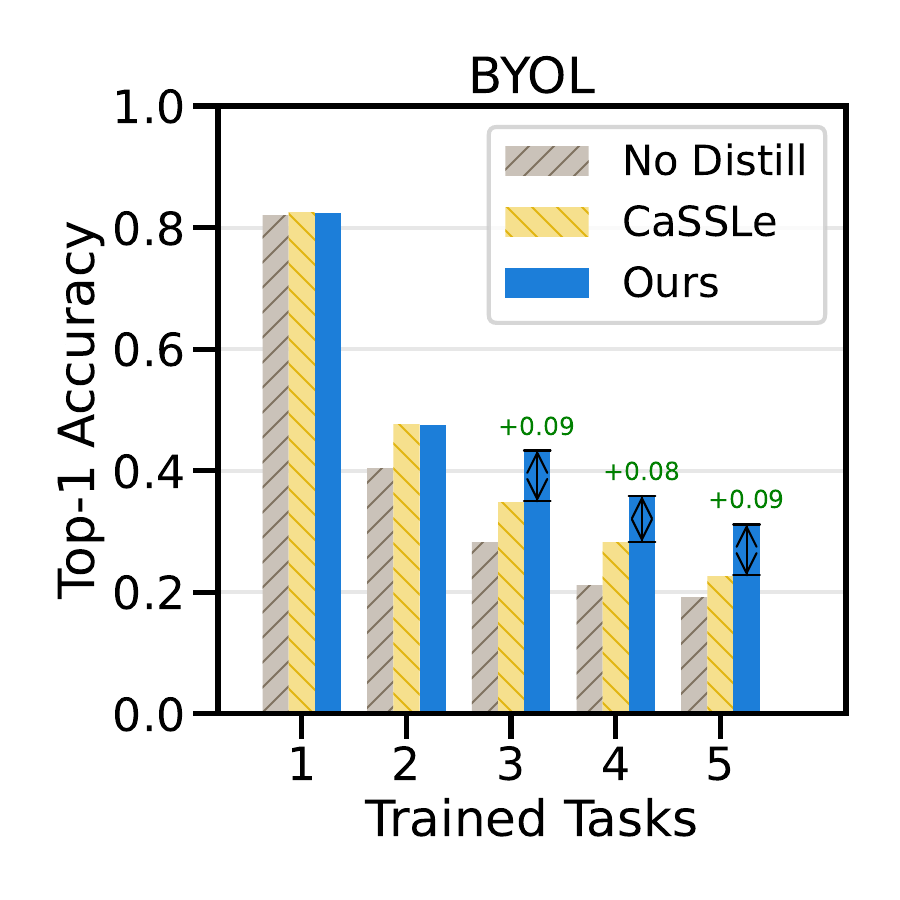}
    \includegraphics[width=0.49 \linewidth ]{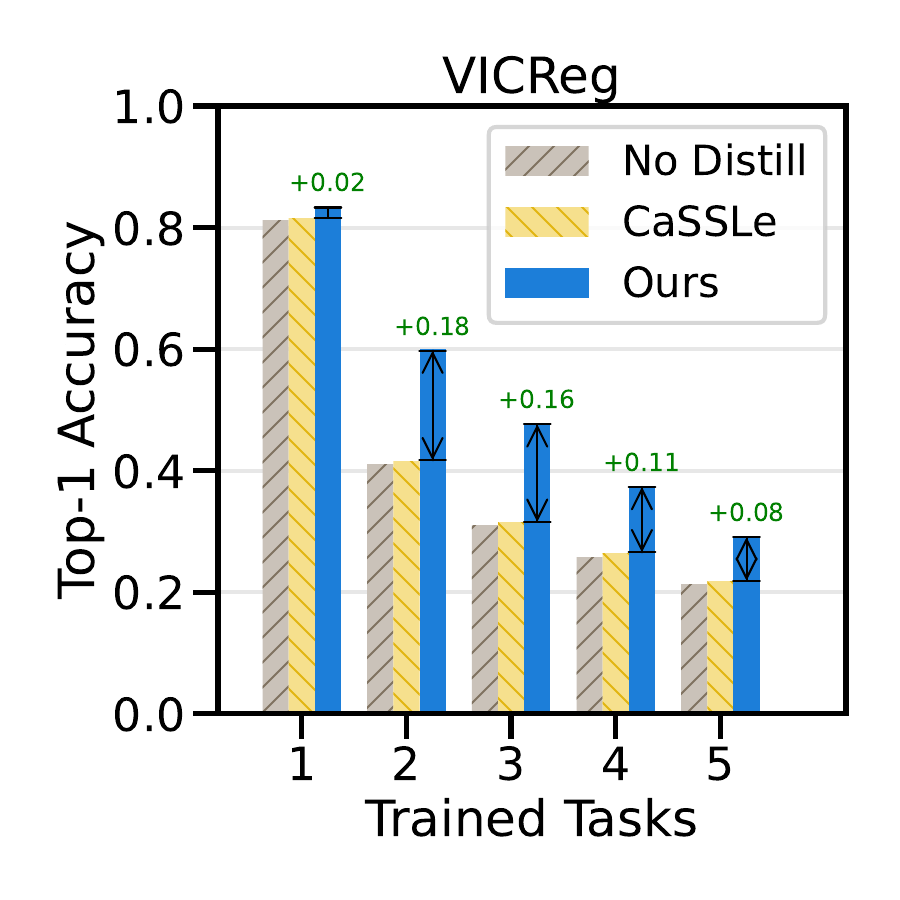}
    \includegraphics[width=0.49 \linewidth ]{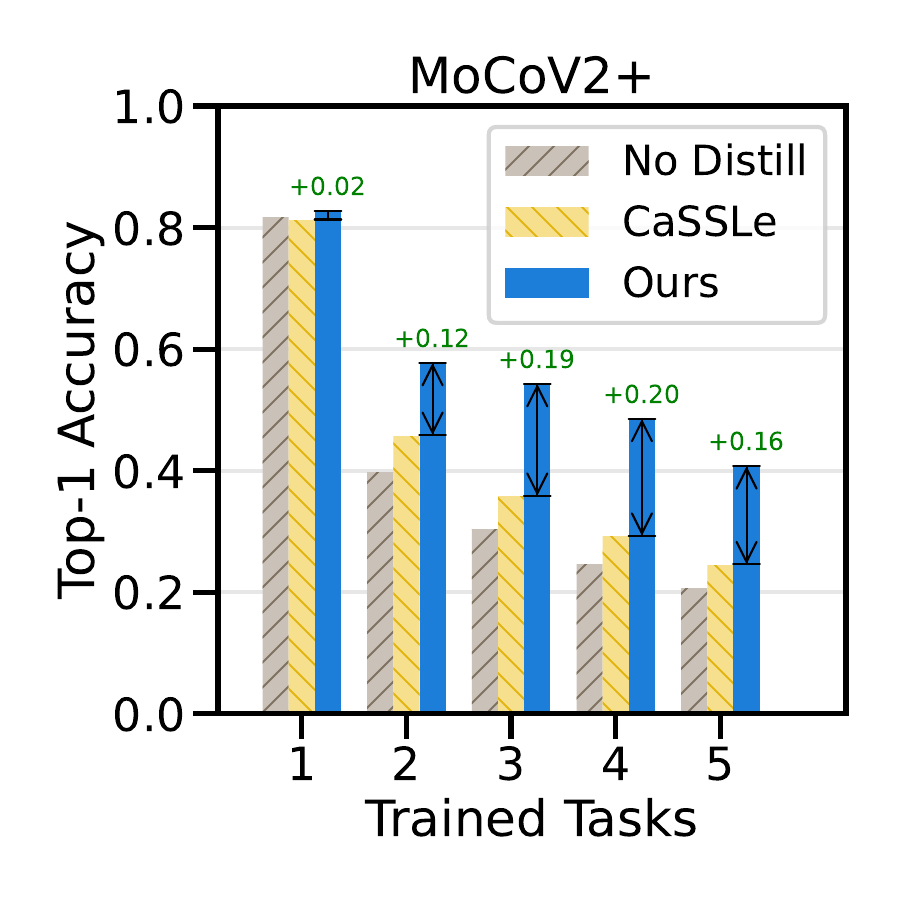}
    \caption{\textbf{Average performance over tasks on CIFAR-100.} Comparison between \tool and baselines using 4 SSL algorithms and 5 tasks. Our model consistently outperforms baselines and is robust to forgetting in later tasks. }
    \label{fig:performance_across_time_cifar}
\end{figure}

\begin{figure}[t]
    \centering
    \includegraphics[width=0.49 \linewidth ]{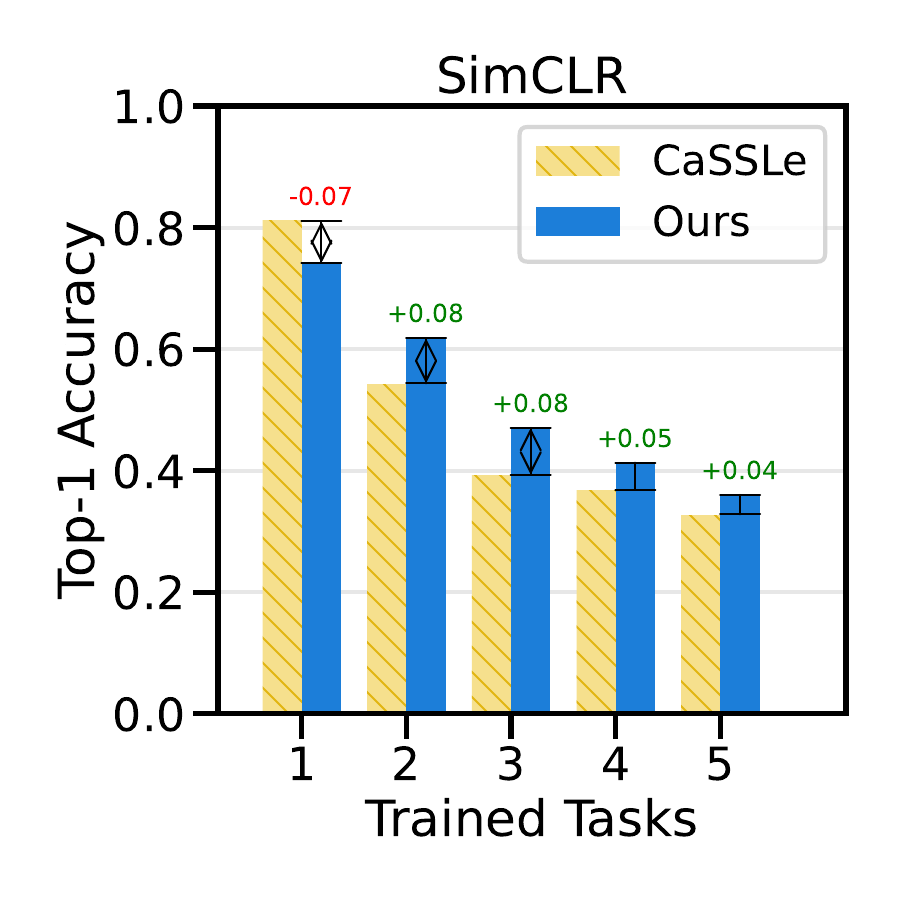}
    \includegraphics[width=0.49 \linewidth ]{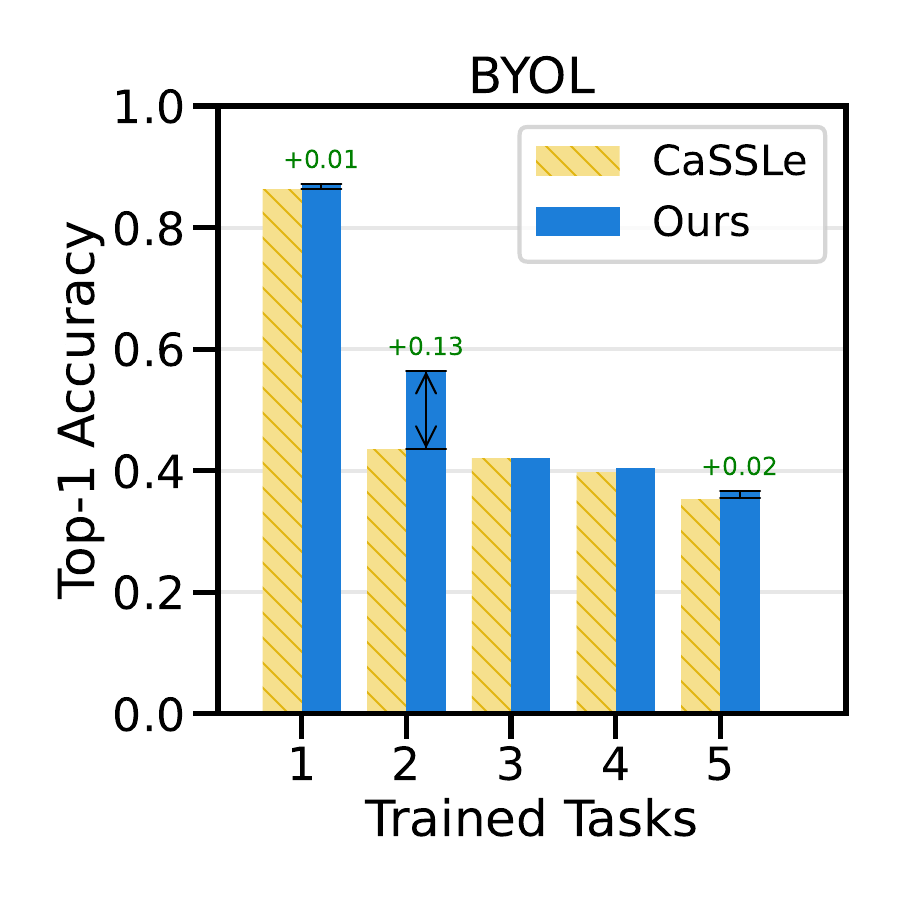}
    \caption{\textbf{Average performance over tasks on ImageNet-100}. Comparison between \tool and CaSSLe for 5 tasks. Our model outperforms in SimCLR and the first tasks in BYOL.}
    \label{fig:performance_across_time_imagenet}
\end{figure}

Here we examine the performance of different methods at different stages of the continual learning process. Figures~\ref{fig:performance_across_time_cifar} and~\ref{fig:performance_across_time_imagenet} show the average accuracy over seen tasks of \tool and the baselines after training on each task, using CIFAR-100 and ImageNet-100 (5 tasks). We did not perform an evaluation on ImageNet-100 for the \emph{No distill} pipeline because it consistently underperforms CaSSLe. We find that in general, all methods show lower performance after training on more tasks, partially due to the fact that the classification problem becomes more difficult as the number of classes increases, as well as catastrophic forgetting. Echoing the results of the previous section, our method maintains a higher level of accuracy overall, even though all methods start from similar performance on the first task. The difference in performance is more evident in CIFAR-100 (Figure~\ref{fig:performance_across_time_cifar}) than in ImageNet-100 (Figure~\ref{fig:performance_across_time_imagenet}), which could be attributed to the larger data size and more available labelled data.

\subsection{Longer continual learning scenarios}

\begin{figure}
    \centering
    \includegraphics[width=\linewidth]{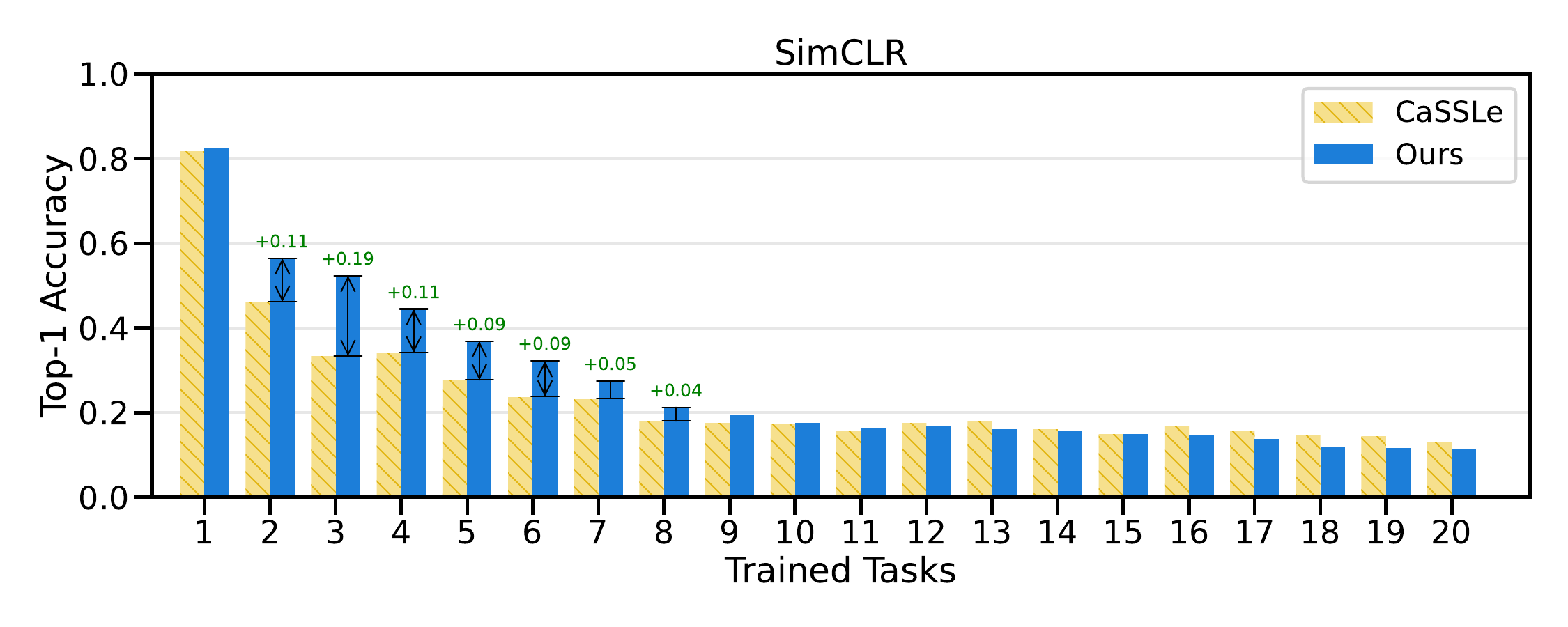}
    \vspace{-0.5cm}
    \caption{\textbf{Average performance over 20 tasks on CIFAR-100.} Our model achieves higher accuracy for the first 10 tasks. }
    \label{fig:performance_20_tasks}
\end{figure}

In certain scenarios, a model could be required to continually learn from data with changing distribution over a long time. Figure~\ref{fig:performance_20_tasks} displays the performance of \tool against CaSSLe on CIFAR-100 when it is split into 20 tasks (5 classes each) instead of 5 tasks only. We can see that \tool is able to outperform the state-of-the-art in the first 10 tasks, but reaches a similar level of performance or even lower in later tasks. We believe that these results demonstrate the trade-off between knowledge retention and learning new tasks: our method trades performance on new tasks for combating catastrophic forgetting. In continual learning with many tasks, this might not be the best strategy because the number of new tasks is usually higher than the number of existing classes. Lower performance on any particular new task might lead to lower performance down the road.

\subsection{Per-task performance breakdown} \label{subsection:per_task}
\begin{figure}[t]
    \centering
    \includegraphics[width=0.49 \linewidth ]{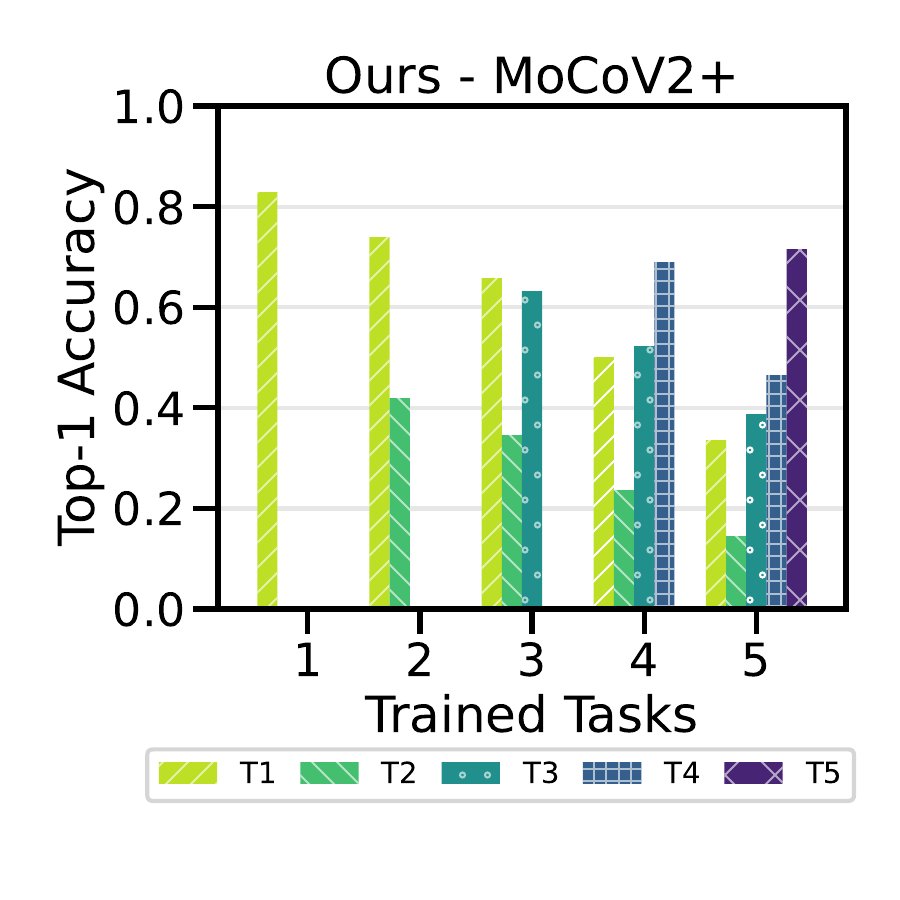}
    \includegraphics[width=0.49 \linewidth ]{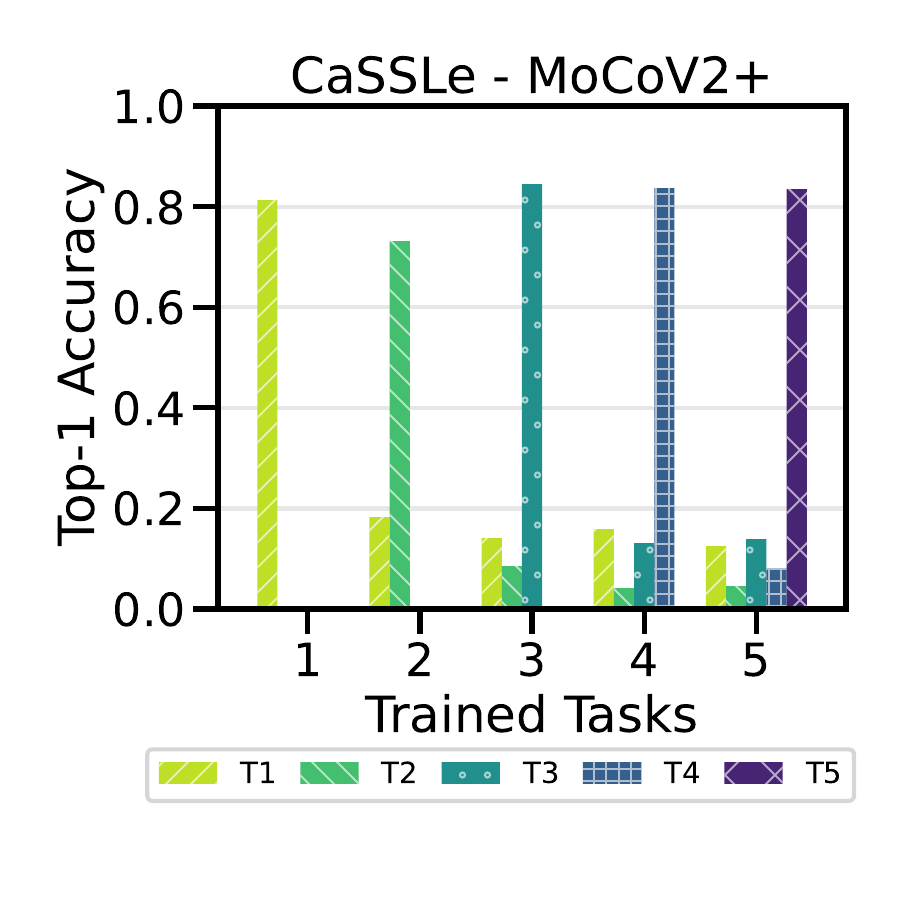}
    \caption{\textbf{Detailed breakdown of performance over tasks on CIFAR-100}. Fine-grained accuracy for every additional task across \tool and CaSSLe, with a fixed SSL backbone.}
    \label{fig:performance_per_task}
\end{figure}

To support our intuition that our method prioritises knowledge retention over learning from new tasks, we investigate the per-task performance at each step. Figure~\ref{fig:performance_per_task} shows the performance of the model on each individual task throughout the continual learning process, using CIFAR-100 with 5 tasks. We find that \tool is able to forget acquired knowledge more gracefully \cite{de2021continual} than CaSSLe, where the performance on previous tasks gracefully degrades over time. Without a knowledge distillation scheme, CaSSLe, on the other hand, is able to perform on the new task better than \tool, but the performance on previous tasks drops significantly in just one step. This shows that our method strikes a balance between performance on new tasks and knowledge retention. The supplementary material contains additional results with ImageNet-100 and other SSL techniques. Trade-off along this spectrum is left for future work.

\subsection{Comprehensive evaluation of continual learning and SSL methods}
\begin{table}
\begin{center}
\footnotesize
\begin{tabularx}{\linewidth}{*{6}l}
\toprule
\textbf{SSL}        &     \textbf{Baseline}     & \textbf{FA} $\uparrow$ & \textbf{CA} $\uparrow$ & \textbf{F} $\downarrow$ & \textbf{FT} $\uparrow$ \\ \midrule
\multirow{3}{*}{BYOL}    & No Distill & 0.191          & 0.381              & 0.794      & \phantom{-}0.003            \\
                         & CaSSLe     & 0.226          & 0.431              & 0.758      & \textbf{\phantom{-}0.010}            \\
                         & \textbf{\tool}       & \textbf{0.313}          & \textbf{0.481}              & \textbf{0.610}      & -0.028           \\ \midrule
\multirow{3}{*}{SimCLR}  & No Distill & 0.209          & 0.391              & 0.755      & \textbf{\phantom{-}\underline{0.021}}          \\
                         & CaSSLe    & 0.213          & 0.403              & 0.747      & \phantom{-}0.019            \\
                         & \textbf{\tool}       & \textbf{0.362}          & \textbf{0.547}              & \textbf{0.451}      & -0.170           \\ \midrule
\multirow{3}{*}{MoCoV2+} & No Distill & 0.206          & 0.393              & 0.760      & \textbf{-0.010}           \\
                         & CaSSLe     & 0.244          & 0.432              & 0.709      & -0.012           \\
                         & \textbf{\tool}       & \textbf{\underline{0.409}}          & \textbf{\underline{0.570}}              & \textbf{\underline{0.396}}     & -0.210          \\ \midrule
\multirow{3}{*}{VICReg}  & No Distill & 0.212          & 0.400              & 0.750      & \textbf{\phantom{-}0.004}            \\
                         & CaSSLe     & 0.217          & 0.405              & 0.739      & -0.001           \\
                         & \textbf{\tool}       & \textbf{0.292}          & \textbf{0.516}              & \textbf{0.580}      & -0.110          \\
\bottomrule
\end{tabularx}
\end{center}
\caption{\textbf{SSL \& baseline performance comparison.} Evaluation of \tool compared to four SSL and two continual baselines across four metrics on CIFAR-100. Our model with the MoCoV2+ backbone outperforms in most metrics. Best performance across SSL methods is bold, and across metrics is underlined. (FA: Final Accuracy, CA: Continual Accuracy, F: Forgetting, FT: Forward Transfer) }
\label{tab:ssl_baseline}
\end{table}
Table~\ref{tab:ssl_baseline} presents \tool compared to other baselines on the four considered SSL paradigms. These results on CIFAR-100 show that \tool improves the performance of all SSL methods as represented by the final and continual accuracy. 
As expected, the absence of distillation in No Distill definitely hurts the accuracy of the overall continual learning framework as it is unable to maintain knowledge with each additional task.  This can also be seen in the Forgetting metric where No Distill is on average 0.255 higher than \tool and 0.026 higher than CaSSLe. \tool always outperforms CaSSLe across the FA, CA, and F metrics achieving the best performance with MoCoV2+. On the other hand, as shown in the results above, CaSSLE and \emph{No distill}, which prioritise learning from new data, show a slight positive transfer in some cases, while our model suffers from negative forward transfer. 

\subsection{Ablation on replay dataset size}
\begin{figure}[t]
    \centering
    \includegraphics[width=0.49 \linewidth ]{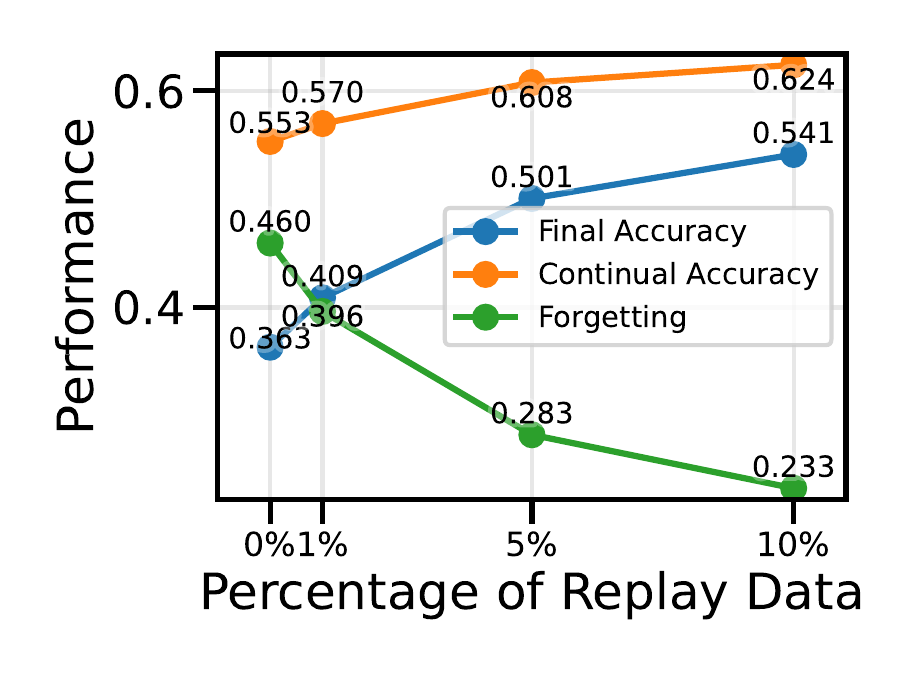}
    \includegraphics[width=0.49 \linewidth ]{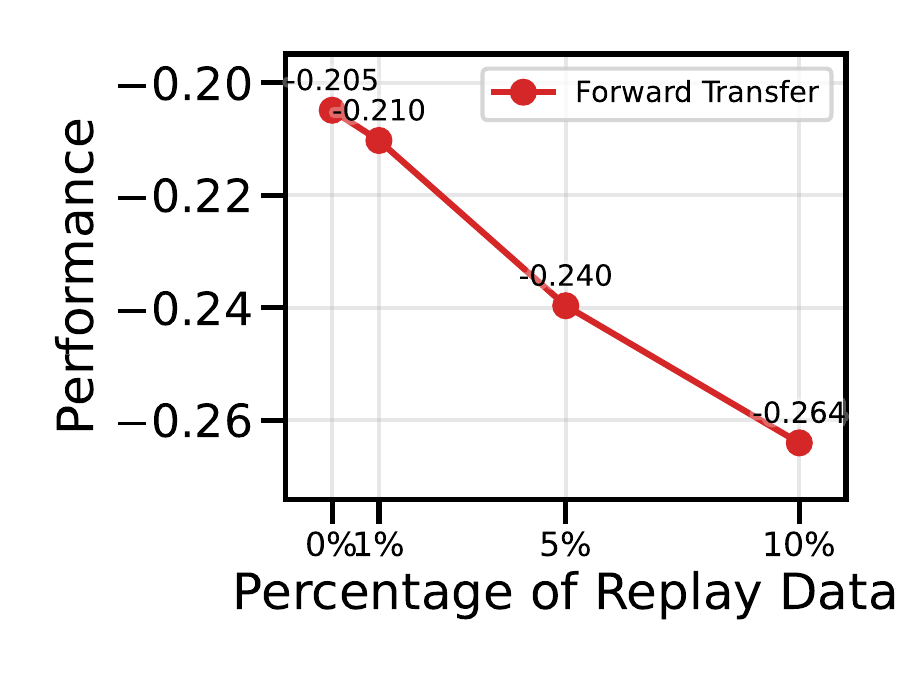}
    \caption{\textbf{Ablation analysis for different amounts of replay data}. Performance of our model on CIFAR100 with MoCoV2+ backbone with varying amounts of replay data (1\% is the default replay in the rest of the paper). A higher amount of replay data leads to better overall performance, less forgetting, but  lower forward transfer.}
    \label{fig:performance_replay}
\end{figure}

Replay was shown to be one of the crucial components in combating catastrophic forgetting for methods that do not rely on task labels \cite{de2021continual}. In this evaluation, we vary the amount of data being replayed to \tool throughout the training process ranging from 0\%, 1\% (default setting in all experiments), 5\% to 10\% on CIFAR-100 using MoCoV2+ as the SSL method, and display these results in Figure~\ref{fig:performance_replay}. As expected, by replaying more data, the model achieves higher continual accuracy as well as higher final accuracy and lower forgetting. However, we cannot see the full picture without looking at the change in forward transfer. Here, we observe that by replaying more data, the model prioritises knowledge retention even more, and forward transfer suffers as a result. This trade-off is important to be taken into account when designing a continual learning system, in addition to storage and privacy concerns. It is important to note that \tool works reasonably well without any replay (0\% case), which would satisfy a strict data assumption when replay is not practical.

\section{Conclusion}
In order to deploy continual learning systems in real-world scenarios, we have to address many challenges that are present in existing works. Towards this goal, here we introduced \tool, a continual learning method that addresses shortcomings of self-supervised and supervised continual learning methods, in which a classifier is continually trained with the possibility of deployment at any point during the training process. We carefully design the learning objective to balance feature extractor and classifier training, as well as knowledge retention and learning from new data. Through extensive evaluation and a comprehensive set of evaluation metrics, we have demonstrated that \tool is able to balance the trade-off between knowledge retention and learning from new data more gracefully compared to the state-of-the-art, and achieves higher performance overall.

\section{Acknowledgements}
This work is partially supported by Nokia Bell Labs through their donation for the Centre of Mobile, Wearable Systems and Augmented Intelligence to the University of Cambridge.

{\small
\bibliographystyle{ieee_fullname}
\bibliography{ref}
}

\clearpage \newpage
\appendix

\begin{appendix}
\section{Additional results}

\begin{figure}
    \centering
    \includegraphics[width=0.49 \linewidth ]{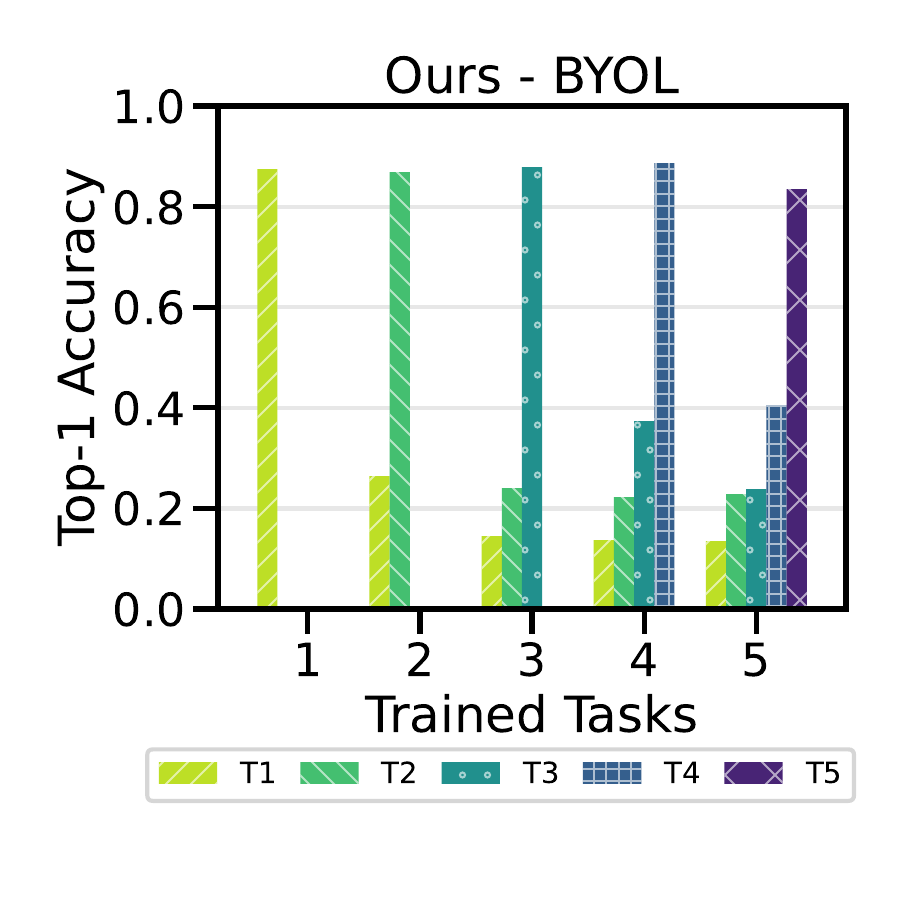}
    \includegraphics[width=0.49 \linewidth ]{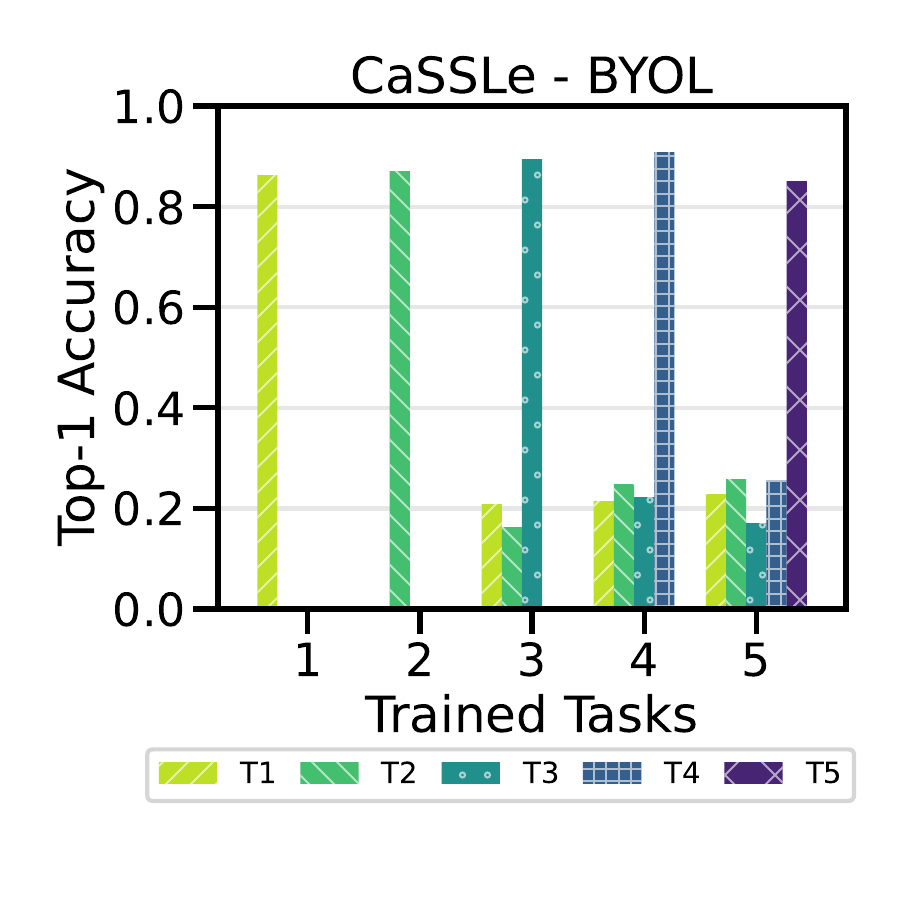}
    \includegraphics[width=0.49 \linewidth ]{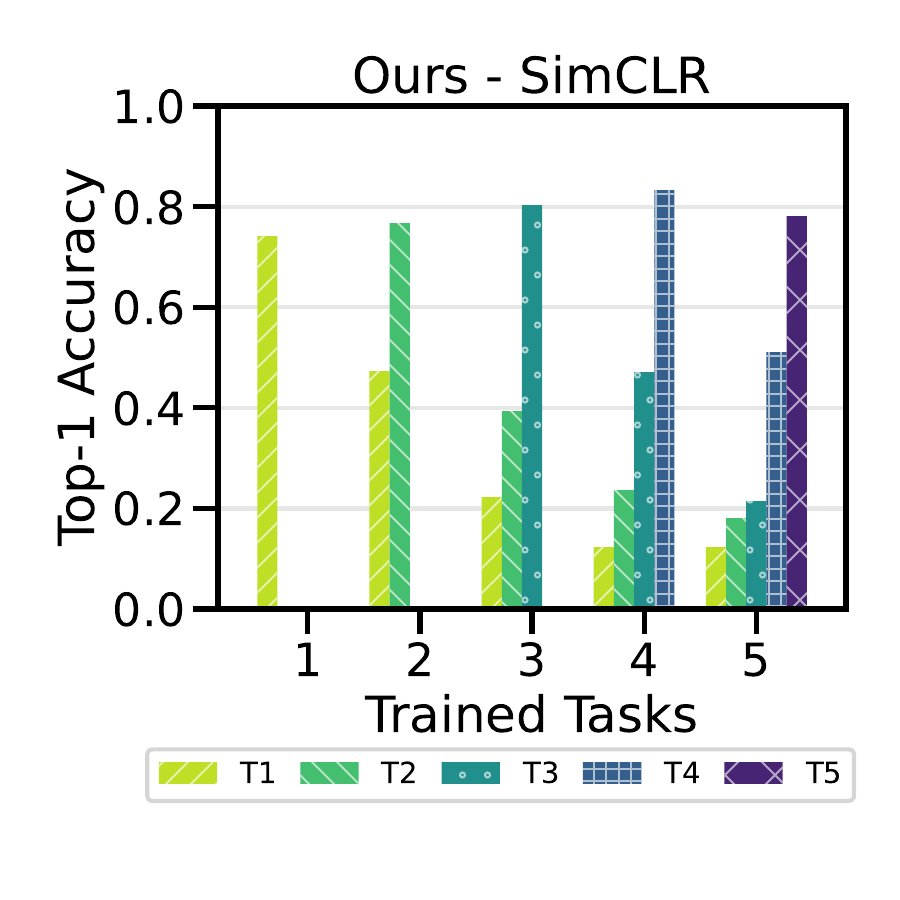}
    \includegraphics[width=0.49 \linewidth ]{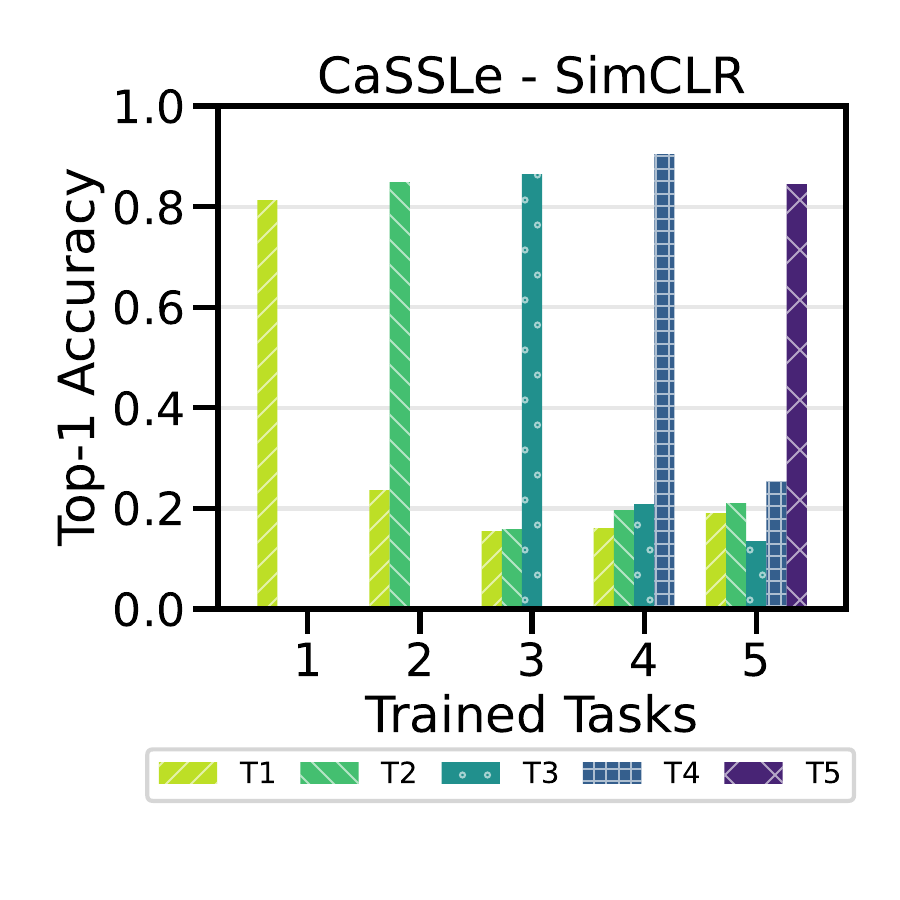}
    \caption{\textbf{Detailed breakdown of performance over tasks on Imagenet-100}. Fine-grained accuracy for every additional task across \tool and CaSSLe, with a fixed SSL backbone.}
    \label{fig:performance_per_task_appendix_imagenet}
\end{figure}

\begin{figure*}
    \centering
    \includegraphics[width=0.3 \linewidth ]{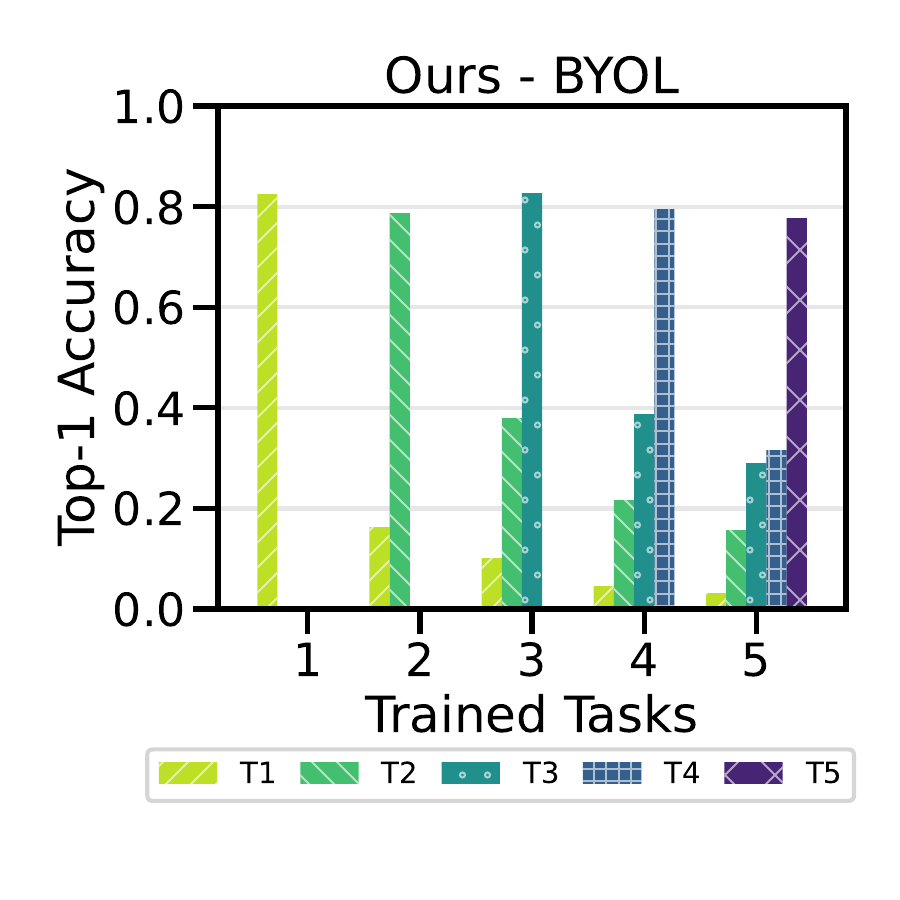}
    \includegraphics[width=0.3 \linewidth ]{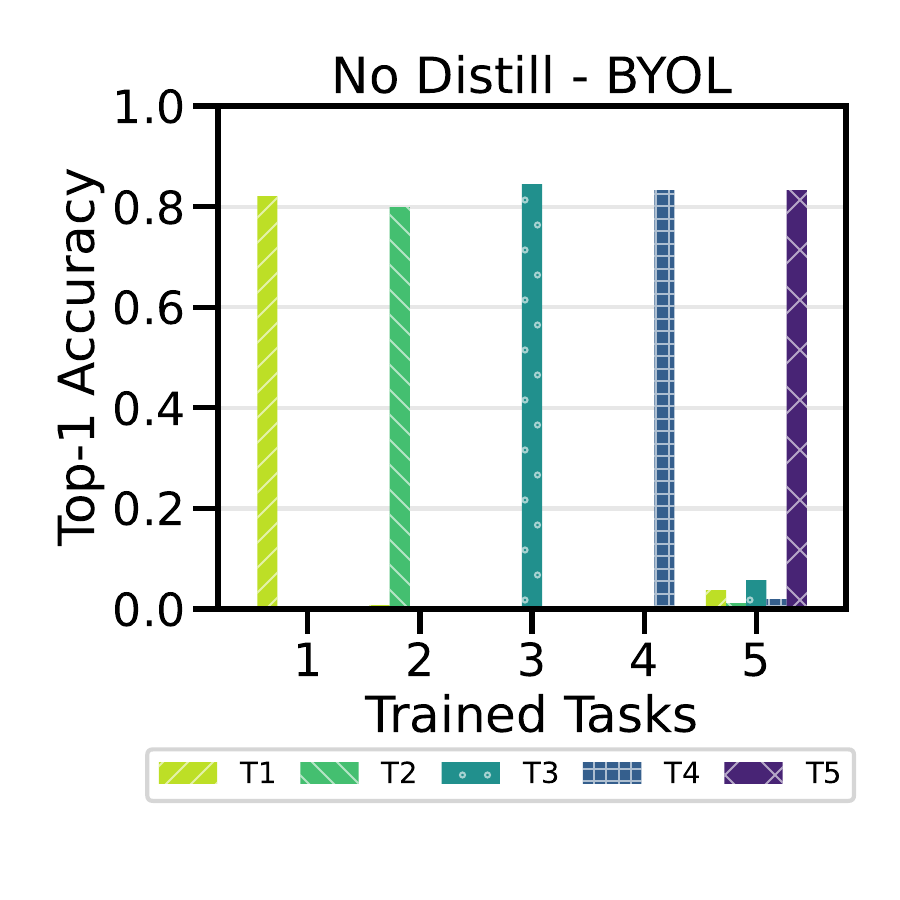}
    \includegraphics[width=0.3 \linewidth ]{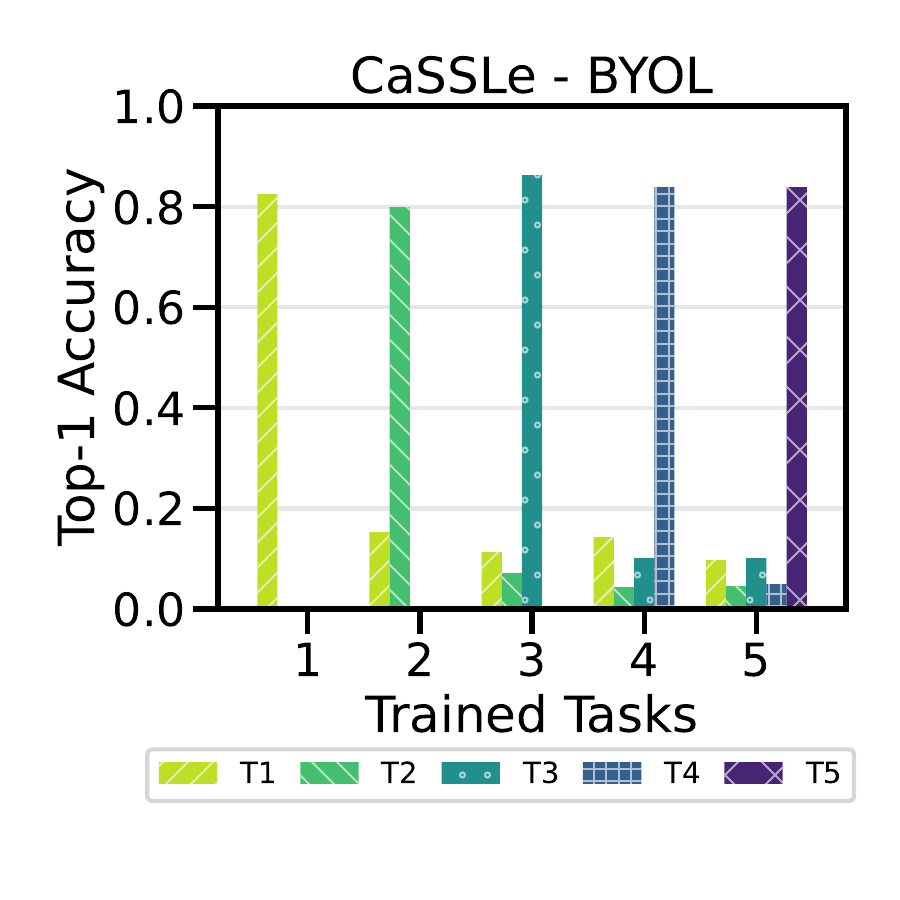}
    \includegraphics[width=0.3 \linewidth ]{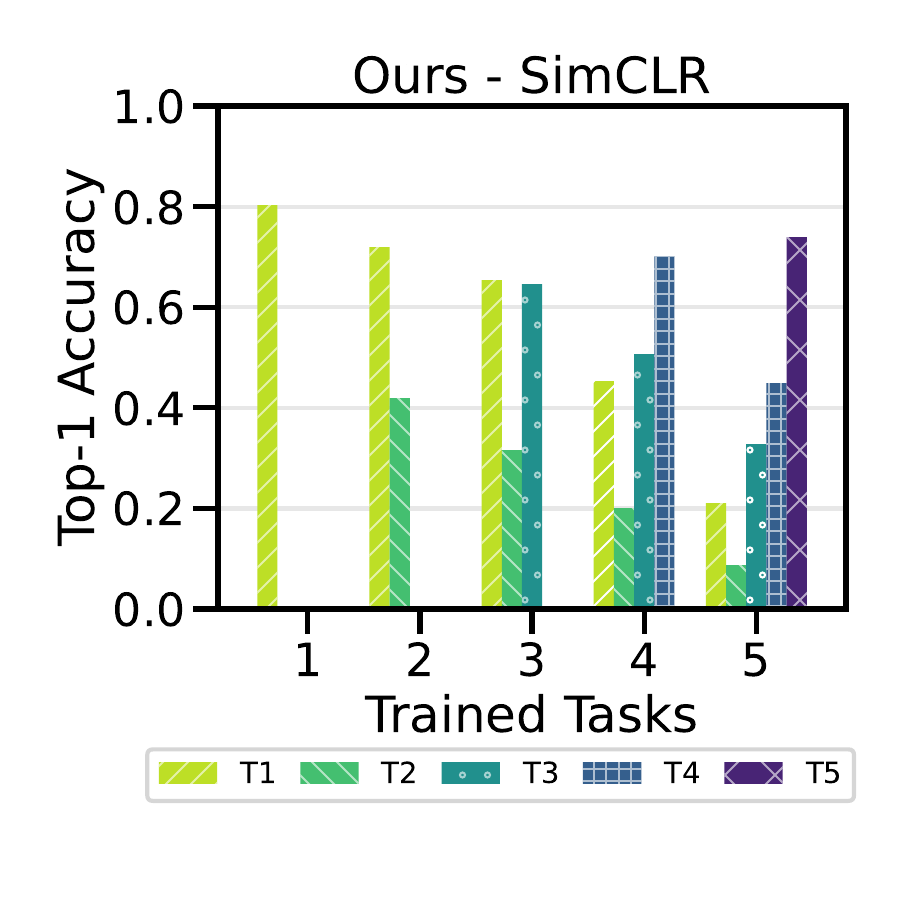}
    \includegraphics[width=0.3 \linewidth ]{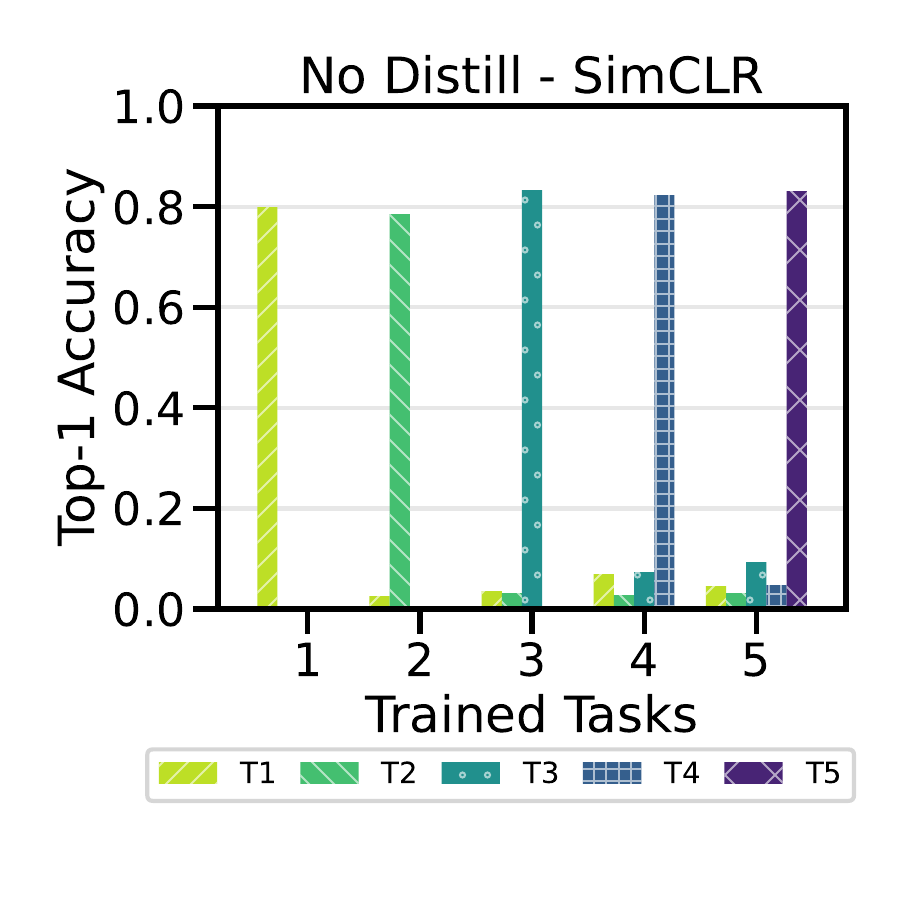}
    \includegraphics[width=0.3 \linewidth ]{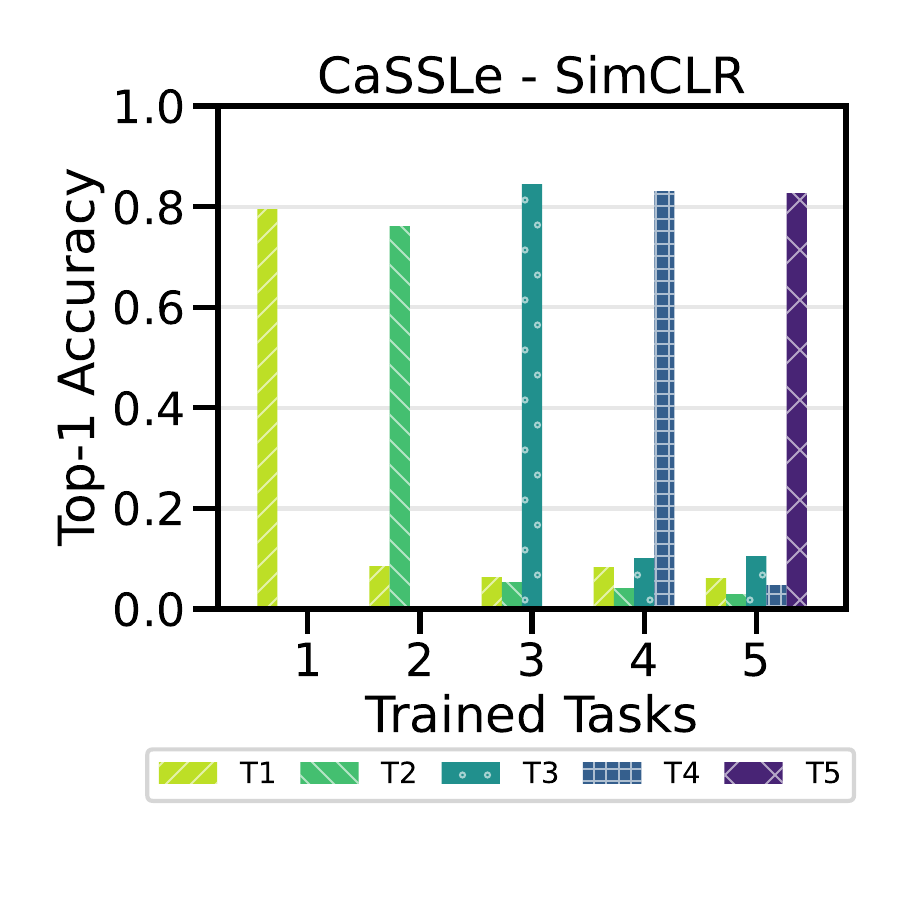}
    \includegraphics[width=0.3 \linewidth ]{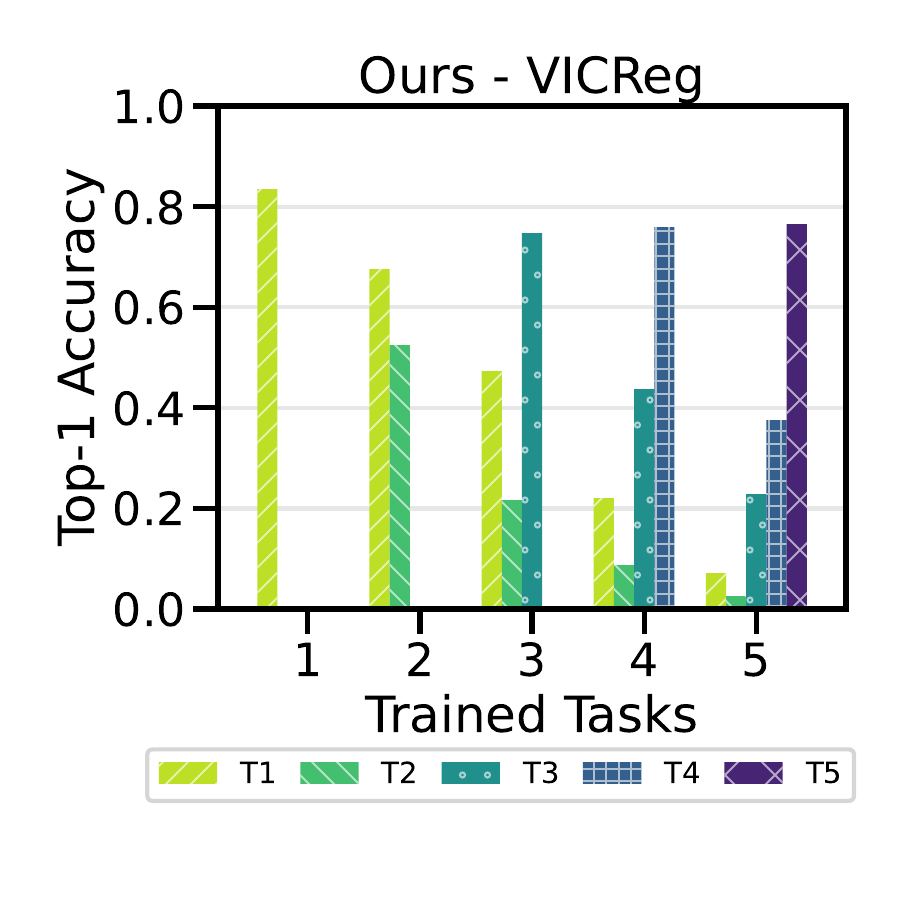}
    \includegraphics[width=0.3 \linewidth ]{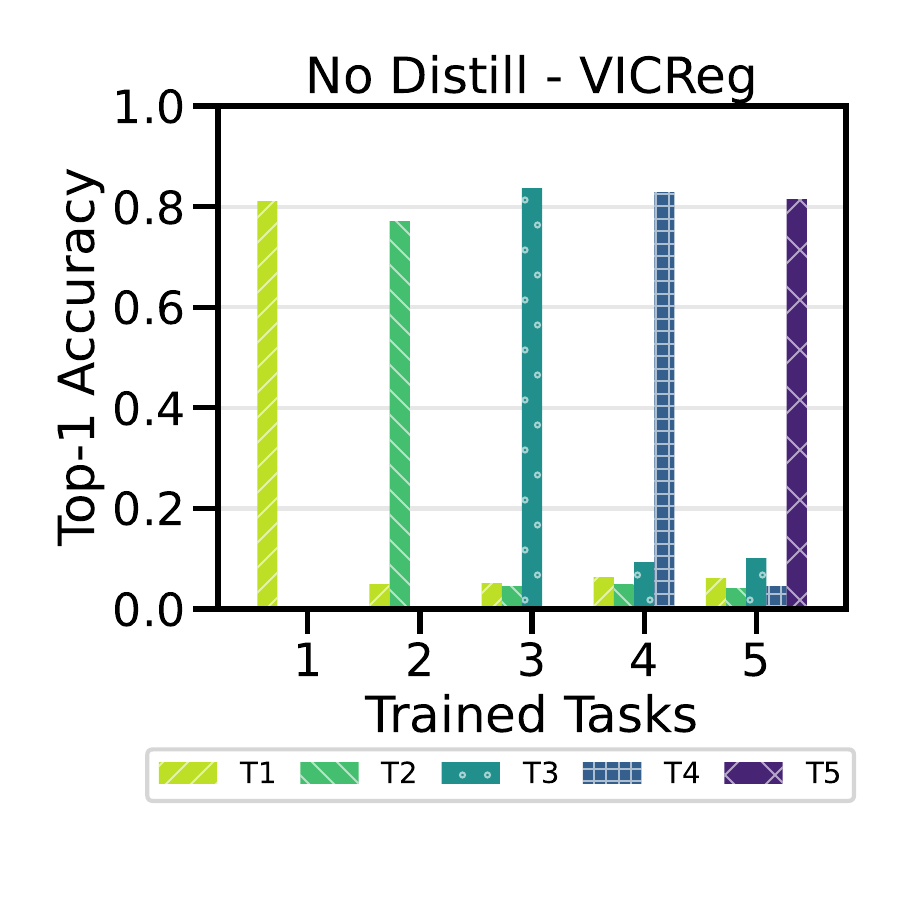}
    \includegraphics[width=0.3 \linewidth ]{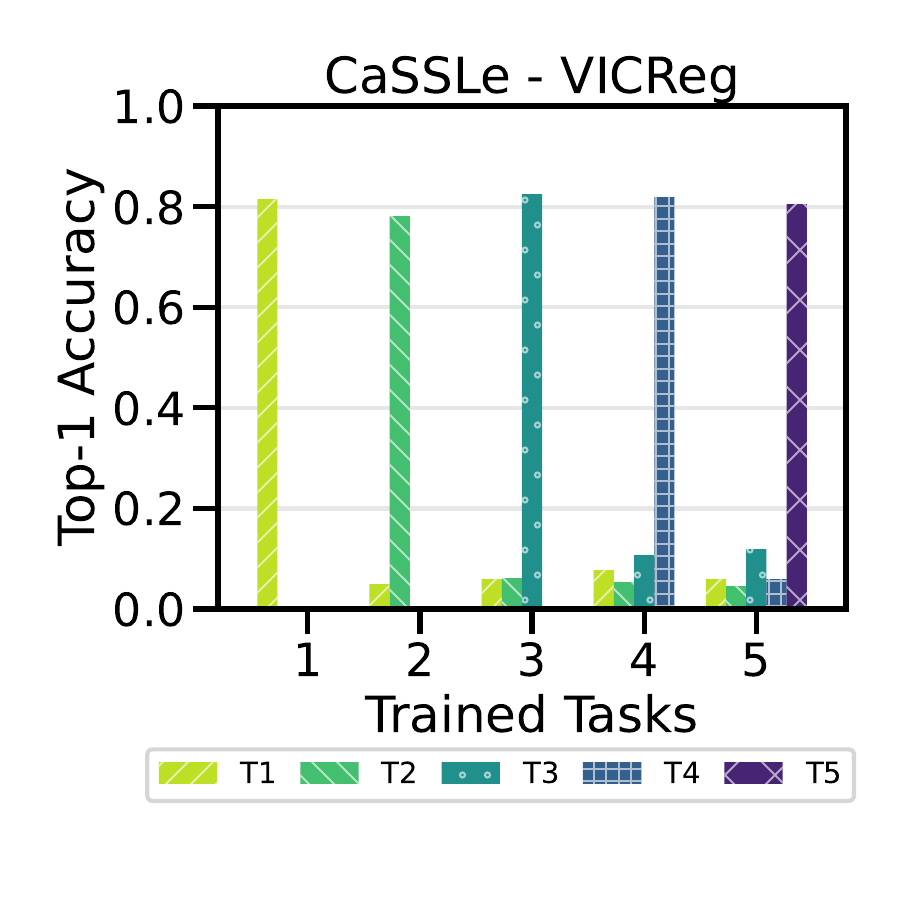}
    \caption{\textbf{Detailed breakdown of performance over tasks on CIFAR-100}. Fine-grained accuracy for every additional task across \tool and CaSSLe, with a fixed SSL backbone.}
    \label{fig:performance_per_task_appendix_cifar}
\end{figure*}

In addition to the results discussed in section \ref{subsection:per_task}, Figure~\ref{fig:performance_per_task_appendix_cifar} displays the performance of the models on each individual task in the continual learning process on CIFAR-100, using different self-supervised learning methods (BYOL~\cite{grill2020bootstrap}, SimCLR~\cite{chen2020simple}, ViCReg~\cite{bardes2022vicreg}) for training and knowledge distillation, and different distillation strategies (\tool, \emph{No Distill} and CaSSLe). We can again observe that our proposal is able to retain knowledge better compared to other methods. The \emph{No Distill} retains the least amount of knowledge where the performance drops to almost zero after learning a new task. We can make similar observations on ImageNet-100 (see Figure~\ref{fig:performance_per_task_appendix_imagenet}), where our method is generally more able to mitigate forgetting, although this effect is more apparent when using SimCLR than BYOL.

\section{Hyperparameters}
In our experiments, we retained the hyperparameters for each self-supervised learning method (BYOL~\cite{grill2020bootstrap}, SimCLR~\cite{chen2020simple}, ViCReg~\cite{bardes2022vicreg}), MoCoV2+~\cite{chen2020improved}) as they were originally proposed in their corresponding works. This was shown to reduce interference~\cite{fini2022self} and used to make sure that the comparison is not a result of hyperparameter tuning. In the loss function (Equation~\ref{eq:loss}), it is possible to use a different weight for each loss function term, to enhance or reduce the supervisory signal from different objectives. A weighting factor of 2 was empirically chosen for the
knowledge distillation loss for the classifier while others are kept as 1. Since the replay dataset is much smaller in size compared to the dataset of the current task, we ensure that each batch that the models are trained on contains at least 32 samples from the replay dataset. The replay dataset is reset as many times as necessary in order to match the number of batches in the data of the current task.

\section{Algorithm}
The algorithm of our training pipeline is provided in this section (Algorithm \ref{algo:kaizen}) to accompany the descriptions given in Section~\ref{subsection:method} and illustrated in Figure~\ref{fig:overview} using a PyTorch-like syntax, which outlines the implementation of \tool.

\begin{algorithm}[t]
   \caption{Algorithm of \tool{}.}
   \label{algo:kaizen}
    \definecolor{codeblue}{rgb}{0.25,0.5,0.5}
    \definecolor{codekw}{rgb}{0.85, 0.18, 0.50}
    \lstset{
      basicstyle=\fontsize{7.2pt}{7.2pt}\ttfamily,
      commentstyle=\fontsize{7.2pt}{7.2pt}\color{codeblue},
      keywordstyle=\fontsize{7.2pt}{7.2pt}\color{codekw},
    }
\begin{lstlisting}[language=python,breaklines]
# aug_f: stochastic augmentation function
# f_o: Current Feature Extractor
# f_t: Momentum Feature Extractor (if applicable)
# f_p: Previous Feature Extractor
# h_o: Predictor for Knowledge Distillation
# h_t: Predictor for SSL
# g_t: Current Classifier
# g_p: Previous Classifier
# loss_ssl: self-supervised learning loss
# loss_ce: cross-entropy loss

def train_step(x, y):
    # augmented views of input
    x1, x2 = aug_f(x), aug_f(x)

    # pass through feature extractors
    z_o = f_o(x1)
    z_t = f_t(x2)
    z_p = f_p(x1)

    # pass embeddings through classifiers
    c_t = g_t(z_t.detach()) # detach stops gradients backpropagation
    c_p = g_p(z_p)

    # pass embeddings through predictors
    p_o = h_o(z_o)
    p_t = h_t(z_o)

    # knowledge distillation for feature extractor
    kd_fe = loss_ssl(p_o, z_p.detach())

    # knowledge distillation for classifier
    kd_c = loss_ce(c_t, c_p.detach())

    # supervised training for current task
    ct_c = loss_ce(c_t, y)

    # SSL training
    ct_fe = loss_ssl(p_t, z_t)

    # Overall loss
    loss = kd_fe + kd_c + ct_c + ct_fe
    
    return loss
\end{lstlisting}
\end{algorithm}

\end{appendix}

\end{document}